\documentclass[journal]{IEEEtran}

\usepackage{amsmath,amsfonts,bm}

\def\eqref#1{equation~\ref{#1}}

\def\1{\bm{1}}

\def\vc{{\bm{c}}}

\def\vg{{\bm{g}}}
\def\vh{{\bm{h}}}

\def\vx{{\bm{x}}}
\def\vy{{\bm{y}}}

\def\mI{{\bm{I}}}

\def\mM{{\bm{M}}}

\def\mP{{\bm{P}}}
\def\mQ{{\bm{Q}}}

\def\mU{{\bm{U}}}

\def\mX{{\bm{X}}}
\def\mY{{\bm{Y}}}

\DeclareMathAlphabet{\mathsfit}{\encodingdefault}{\sfdefault}{m}{sl}
\SetMathAlphabet{\mathsfit}{bold}{\encodingdefault}{\sfdefault}{bx}{n}

\def\gD{{\mathcal{D}}}
\def\gE{{\mathcal{E}}}

\def\gG{{\mathcal{G}}}
\def\gH{{\mathcal{H}}}

\def\gL{{\mathcal{L}}}
\def\gM{{\mathcal{M}}}
\def\gN{{\mathcal{N}}}

\def\gQ{{\mathcal{Q}}}
\def\gR{{\mathcal{R}}}
\def\gS{{\mathcal{S}}}
\def\gT{{\mathcal{T}}}

\def\sR{{\mathbb{R}}}

\newcommand{\E}{\mathbb{E}}

\usepackage{amsmath,amsopn,amssymb}
\usepackage{graphicx,xspace,color,soul}
\usepackage{epsfig,subfigure}
\usepackage{longtable,multirow}
\usepackage{tabularx}
\usepackage{array,float}
\usepackage{cite}
\usepackage{algorithmic}
\usepackage[linesnumbered, ruled]{algorithm2e}
\usepackage{bm,epstopdf}
\usepackage{caption2}
\usepackage{booktabs}
\newcolumntype{Y}{>{\centering\arraybackslash}X}

\def\ie{{\textit{i.e.}}}
\def\eg{{\textit{e.g.}}}

\def\etal{{\textit{et al.~}}}

\begin{document}

\title{Dynamic Point Cloud Denoising via Gradient Fields}

\author{
	Qianjiang~Hu,
    and~Wei~Hu,~\IEEEmembership{Senior~Member,~IEEE}
	\thanks{Q. Hu and W. Hu are with Wangxuan Institute of Computer Technology, Peking University, No. 128, Zhongguancun North Street, Beijing, China. E-mails: \{hqjpku, forhuwei\}@pku.edu.cn. Corresponding author: Wei Hu. }

}


\maketitle

\begin{abstract}
3D dynamic point clouds provide a discrete representation of real-world objects or scenes in motion, which have been widely applied in immersive telepresence, autonomous driving, surveillance, \textit{etc}.
However, point clouds acquired from sensors are usually perturbed by noise, which affects downstream tasks such as surface reconstruction and analysis.
Although many efforts have been made for static point cloud denoising, dynamic point cloud denoising remains under-explored.
In this paper, we propose a novel gradient-field-based dynamic point cloud denoising method, exploiting the temporal correspondence via the estimation of gradient fields---a fundamental problem in dynamic point cloud processing and analysis.
The gradient field is the gradient of the log-probability function of the noisy point cloud, based on which we perform gradient ascent so as to converge each point to the underlying clean surface. 
We estimate the gradient of each surface patch and exploit the temporal correspondence, where the temporally corresponding patches are searched leveraging on rigid motion in classical mechanics. 
In particular, we treat each patch as a rigid object, which moves in the gradient field of an adjacent frame via force until reaching a balanced state, \ie, when the sum of gradients over the patch reaches 0. 
Since the gradient would be smaller when the point is closer to the underlying surface, the balanced patch would fit the underlying surface well, thus leading to the temporal correspondence. 
Finally, the position of each point in the patch is updated along the direction of the gradient averaged from corresponding patches in adjacent frames.
Experimental results demonstrate that the proposed model outperforms state-of-the-art methods under both synthetic noise and simulated real-world noise.       

\end{abstract}


\section{Introduction}
\label{sec:intro}
\begin{figure}[t]
\begin{center}
    \includegraphics[width=\columnwidth]{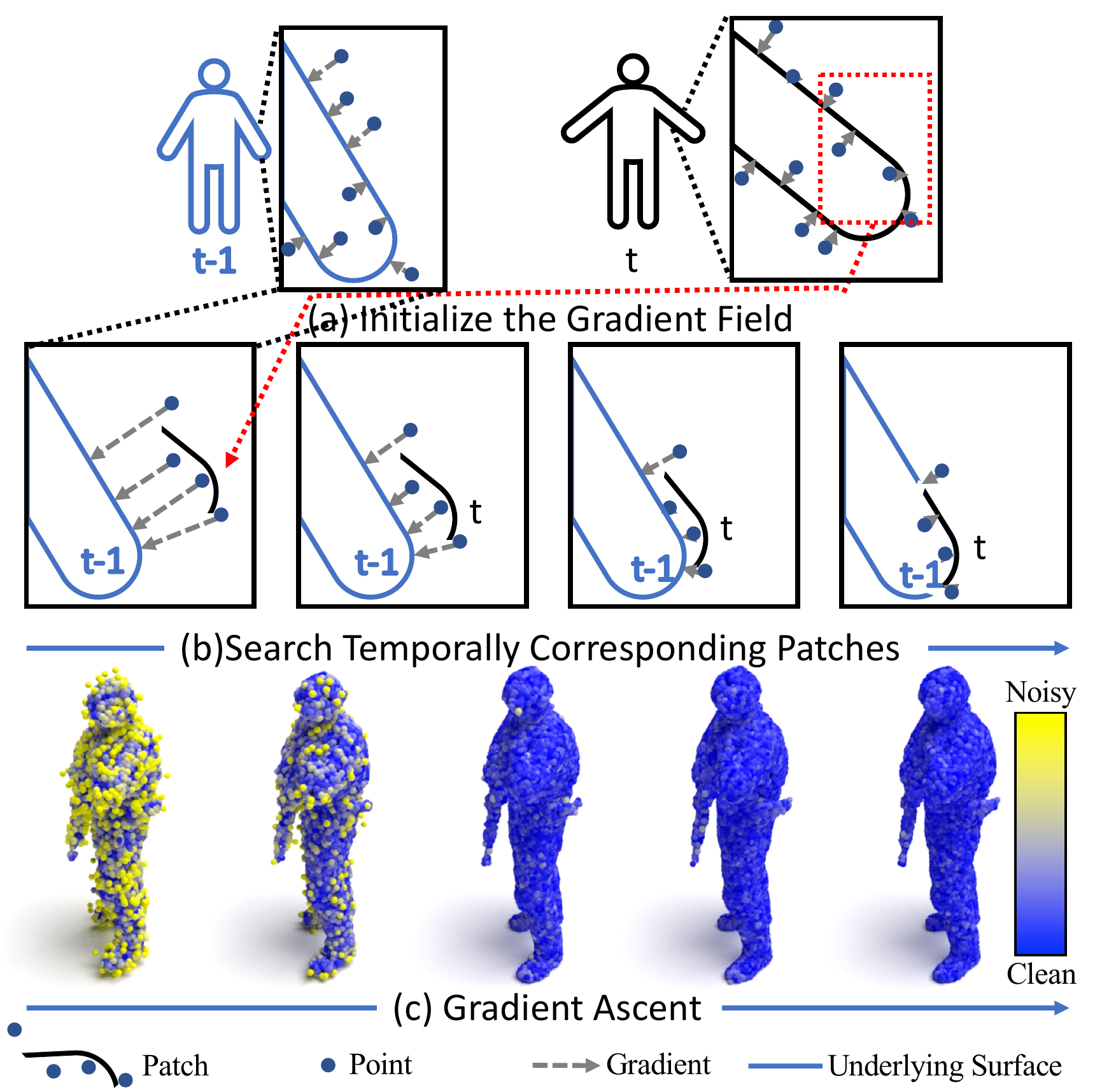}
\end{center}
\vspace{-0.1in}
  \caption{Illustration of the proposed dynamic point cloud denoising method. \textbf{Top:} We first estimate the initial gradient field in each static point cloud frame. \textbf{Middle:} Then we search temporally corresponding patches across adjacent frames in the gradient field. \textbf{Bottom:} Finally, we take the average of the gradient fields of corresponding patches to acquire the final gradient field, based on which we perform gradient ascent iteratively to converge noisy points to the underlying surface for denoising.}
\label{fig:teaser}
\end{figure}

The maturity of laser scanning has enabled the convenient acquisition of 3D dynamic point clouds---a discrete representation of 3D objects or scenes in motion, which have various applications in autonomous driving, robotics, and immersive tele-presence.
A dynamic point cloud consists of a sequence of static point clouds, each of which is composed of a set of points irregularly sampled from the continuous surfaces of objects or scenes.
However, 3D dynamic point clouds often suffer from noise due to the inherent limitations of acquisition equipments or computational errors of sampling algorithms, which affects downstream tasks such as recognition and analysis.
Hence, 3D dynamic point cloud denoising is crucial to relevant 3D applications.

Recently, numerous approaches have been proposed for static point cloud denoising, which can be categorized into two major classes: optimization-based methods and deep-learning-based methods.
Optimization-based methods usually denoise point clouds with optimization algorithms \cite{Alexa2003Computing,A2009Feature,Lipman2007Parameterization,Hui2009Consolidation,Huang2013Edge,mattei2017MRPCA,hu2020featuregraph,sch2015graphbased,zeng2019GLR}, which rely heavily on geometric priors and are sometimes difficult to
keep a balance between the denoising effectiveness and detail preservation. 
Thanks to the advent of neural network architectures designed for point clouds \cite{qi2017pointnet, qi2017pointnet2, wang2019dynamic}, deep-learning-based methods \cite{hermosilla2019TotalDenoising,duan2019NeuralProj,rakotosaona2020PCN,pistilli2020learning,luo2020DMR,luo2021score,chen2021deep} have achieved promising denoising performance. 
The majority of deep-learning-based methods predict the displacement of each noisy point from the underlying surface and then apply the inverse displacement to the noisy point clouds \cite{hermosilla2019TotalDenoising,duan2019NeuralProj,rakotosaona2020PCN,pistilli2020learning}, which suffer from shrinkage and outliers due to over-estimation or under-estimation of the displacement.  
Instead, Luo \etal \cite{luo2020DMR} proposed to learn the underlying manifold of a noisy point cloud for reconstruction in a downsample-upsample architecture, which however may cause detail loss during the downsampling stage especially at low noise levels.
Recently, Luo \etal \cite{luo2021score} proposed score-based point cloud denoising, where the position of each point is iteratively updated by gradient ascent from the log-likelihood distribution of the noisy point cloud. 
Although a plethora of approaches have been proposed for static point cloud denoising, few algorithms are designed for dynamic point cloud denoising. 
This problem is also quite challenging, because each point cloud frame is irregularly sampled and different frames may exhibit varying sampling patterns with possibly different numbers of points, which means there is no explicit temporal correspondence between points over time. 

To this end, we propose a gradient-field-based dynamic point cloud denoising paradigm by exploiting the temporal correspondence among adjacent frames, leveraging on the distribution of the noisy point cloud and rigid motion in classical mechanics. 
Dynamic point clouds consist of discrete points $\vx$ sampled from the surface of 3D dynamic objects or scenes, and thus can be modeled as a set of samples from some dynamic 3D distribution $p_t(\vx)$ supported by dynamic 2D manifolds. 
When the dynamic point cloud suffers from noise following some distribution $n$, the distribution of the noisy dynamic point cloud can be modeled as the convolution between the original distribution and noise, expressed as $(p_t * n)(\vx)$. 
Under some mild assumptions of noise $n$ (See Section~\ref{subsec:static_modeling} for details), the mode of $p_t * n$ is the underlying clean surface having higher probability than its ambient space. 
According to this observation, restoring a noisy point cloud naturally amounts to maximizing $(p_t * n)(\vx)$.  
However, the distribution $(p_t * n)(\vx)$ is difficult to acquire in general. 
Instead, we estimate the gradient of the log-probability function, \textit{i.e.}, $\nabla_{\vx}\log[(p_t * n)(\vx)]$, which points to the underlying clean surface (see Fig.~\ref{fig:teaser}(a)). 
Based on the estimated gradient, point cloud denoising can be realized by moving perturbed points towards the underlying surface via gradient ascent.

The key of dynamic point cloud denoising is to search the temporal correspondence among adjacent frames for consistency in the temporal domain. 
Enlightened by free fall in classical mechanics where rigid bodies move spontaneously from places with high energy to places with low potential energy, we design a process that a surface patch of point cloud will spontaneously move to the underlying surface of its adjacent frames to acquire its temporal correspondence.
Inspired by classical mechanics where the negative gradient of the potential energy of a rigid body represents the force that the rigid body receives along the direction of the gradient, we simulate the motion of objects in the gradient field for the search of the temporal correspondence.
If we define the negative log-density $-\log(p*n)$ as the potential energy of a noisy point cloud, then the negative gradient of the potential energy, \ie, $\nabla_\vx\log(p*n)$, could be seen as a force that pushes noisy points towards the underlying surface.
Therefore, we let the patch move in the gradient field of adjacent frames to fit the local structures in consecutive frames until the balanced state is reached, \ie, when the sum of gradients over the patch reaches 0.
Since a point is on the underlying surface when the gradient is 0, the balanced patch would fit the underlying surface well, thus leading to the temporal correspondence.
The gradient of the current surface patch is then estimated as the average of gradients in the temporally corresponding patches in consecutive frames, and the position of each noisy point in the patch is converged to the underlying surface along the direction of the estimated gradient.

Specifically, we first divide each noisy point cloud in the sequence into surface patches, and design a network to estimate the initial gradient field from noisy patches in each frame, which also serves for the subsequent temporal correspondence search.
Next, we treat each noisy patch as a rigid body, and move them in the gradient field of adjacent frames to search for corresponding patches in consecutive frames, which essentially performs rotation and translation over the patch.
Then, we take the average of the gradients of the inverse transformed corresponding patches in consecutive frames as the temporal gradient.
Finally, the dynamic point cloud is denoised by gradient ascent based on the estimated temporal gradient.

The main contributions of this paper are as follows.
\vspace{-1.5mm}
\begin{itemize}
\setlength{\itemsep}{0pt}
\setlength{\parskip}{0pt}
    \item We propose a temporal-gradient learning framework for dynamic point cloud denoising, exploiting the temporal correspondence by leveraging on rigid motion in classical mechanics. 
    
    \item We propose a novel temporal correspondence search approach, which moves each patch in the gradient field by ``force" in classical mechanics until reaching a balanced state to acquire the temporal correspondence. 
    
    
    \item Experimental results demonstrate the effectiveness of the proposed method over the state-of-the-art under both synthetic noise and simulated real-world noise. 
    
\end{itemize}

\section{Related Work}
\label{sec:relatd}

Most existing point cloud denoising methods focus on static point clouds, while dynamic point cloud denoising is under-explored in the literature.
Previous works can be divided into two major classes: \textbf{Optimization-based} methods and \textbf{Deep-learning-based} methods. 

\subsection{Optimization-based Denoising}
Before the advent of deep learning methods, point cloud denoising algorithms were mostly optimization algorithms based on geometric priors. They can be classified into five categories.

\noindent(1) \textbf{Moving-Least-Squares-Based} methods aim to approximate a smooth surface from the input point cloud and minimize the geometric error of the approximation.
\cite{Alexa2003Computing} addressed the denoising problem with a polynomial function in terms of moving least squares (MLS) on a local reference domain to best fit neighboring points.
\cite{A2009Feature} extended it with the strength of robust statistics to reduce the sensitivity to outliers.

\noindent(2) \textbf{Locally-Optimal-Projection-Based} methods also employ the surface approximation for 
denoising point clouds. \cite{Lipman2007Parameterization} proposed locally optimal projection (LOP) that generates a point set to represent the underlying surface while enforcing a uniform distribution over the point cloud.
Based on this method, \cite{Hui2009Consolidation} proposed a weighted locally optimal projection (WLOP) operator to produce a set of denoised, outlier-free and evenly distributed points over the original dense point cloud.
\cite{Huang2013Edge} modified WLOP with an anisotropic weighting function so as to produce point clouds with sharp features preserved. 

\noindent(3) \textbf{Sparsity-Based} methods are based on the theory of sparse representations. 
In sparsity-based methods, a sparse reconstruction of the surface normals is firstly obtained by solving a minimization problem with sparsity regularization, and then point positions are updated by solving another minimization problem based on a local planar assumption.
MRPCA \cite{mattei2017MRPCA} proposed to denoise point cloud patches with a robust PCA algorithm, and then estimated point positions by local averaging.
\cite{arvanitis2018outliers} recognized and removed outliers from a dynamic point cloud sequence using a Robust PCA approach. 

\noindent(4) \textbf{Non-Local-Based} methods exploit non-local similarities among surface patches in a point cloud. 
These methods extend non-local mean (NLM) \cite{buades2005non} and BM3D \cite{dabov2007image} in image denoising to point clouds.  
\cite{digne2012similarity} utilizes the NLM algorithm to denoise point clouds, while \cite{rosman2013patch} proposed to denoise a point cloud by deploying a BM3D algorithm.
With the help of non-local similarities, these methods are able to reduce noise while preserving the surface details.
To calculate the point similarity better, \cite{deschaud10} describes the neighborhood of each point by the polynomial coefficients of the local MLS surface. 
\cite{sarkar2018structured} proposed to smooth point clouds by solving a structured low-rank matrix factorization problem, where a low-rank dictionary representation of patches is optimized.
\cite{zhou2021point} captures non-local similarities by normal height projection, where the most similar non-local projective height vectors are grouped into a height matrix and optimized with an improved weighted nuclear norm minimization.

\noindent(5) \textbf{Graph-Based} methods represent a point cloud over a graph, and essentially perform graph filtering for point cloud denoising \cite{hu2020featuregraph,sch2015graphbased,zeng2019GLR,hu2021graph}.
\cite{zeng20183d} represents point clouds by a low-dimensional manifold model (LDMM), and then performs denoising with graph Laplacian regularization (GLR). \cite{gao2018graph} first removes outliers based on the sparsity of the neighborhood and then removes surface noise with an optimization process regularized by graph-signal smoothness prior. \cite{duan2018weighted} estimates the local tangent plane of each point based on a graph, and then takes the average of the projections of points on multiple tangent planes to reconstruct the point cloud. \cite{hu2020featuregraph} proposed feature graph learning by minimizing the GLR using the Mahalanobis distance metric matrix as a variable, assuming a feature vector per node is available. 
Further, a fast eigen-decomposition-free algorithm is proposed and applied to point cloud denoising, where the graph for each set of self-similar patches is computed from 3D coordinates and surface normals
as features. Based on the assumption that color attributes are correlated with the geometry, \cite{irfan20213d} deploys a graph-based Tikhonov regularization to jointly denoise the geometry and color. 
Similarly, \cite{irfan2021joint} proposed a non-iterative set-up based on spectral graph wavelet transform (SGW) to perform denoising of geometry and color attributes in the graph spectral domain. \cite{hu2021dynamic} represents dynamic point clouds on spatial-temporal graphs, and exploits the temporal consistency via a manifold-to-manifold distance. The problem of dynamic point cloud denoising is then formulated as the joint optimization of the desired point cloud and underlying spatial-temporal graph representation.


\subsection{Deep-learning-based Denoising}
Benefiting from the maturity of point-based neural networks \cite{qi2017pointnet,qi2017pointnet2,wang2019dynamic}, many deep point cloud denoising approaches have been proposed.

\noindent(1) \textbf{Displacement-Based} methods employ a neural network to predict the displacement of each point in a noisy point cloud, and apply the inverse displacement to each point. 
PointCleanNet (PCNet) \cite{rakotosaona2020PCN} is the pioneer of this class of approaches, which employs a variant of PointNet as its backbone network.
With graph convolutional networks (GCN), \cite{pistilli2020learning} proposed GPDNet to enhance the robustness of the neural denoiser.
Assuming that points with denser surroundings are closer to the underlying surface, Total Denoising (TotalDn) \cite{hermosilla2019TotalDenoising} proposed an unsupervised loss function to adapt to unsupervised scenarios. 
However, displacement-based methods generally suffer from shrinkage and outliers, as a result of inaccurate estimation of the noise displacement.

\noindent(2) \textbf{Downsampling-Upsampling} methods such as \cite{luo2020DMR} learn the underlying manifold of a noisy point cloud from subsampled points, and resample on the reconstructed manifold to obtain the noise-free point cloud. 
Although the downsampling stage discards outliers in the input, it may also discard some informative details, resulting in over-smoothing especially at low noise levels.

\noindent(3) \textbf{Score-Based (Gradient-based)} methods deploy a score matching technique to denoise point clouds. Score matching is a technique for training energy-based models, which belongs to the family of non-normalized probability distributions \cite{lecun2006tutorial}.
It deals with matching the model-predicted gradients and the data log-density gradients by minimizing the squared distance between them \cite{hyvarinen2005estimation, song2019generative}. 
In particular, \cite{luo2021score} estimates the gradient of the noisy point cloud and performs denoising via gradient ascent. 
The extended work \cite{chen2021deep} proposed a global and continuous gradient field model, and resampled degraded point clouds via gradient ascent with the Graph Laplacian Regularizer introduced. 

Our method belongs to this category. 
Different from \cite{luo2021score}, we focus on the much more challenging dynamic point cloud denoising, and the proposed method distinguishes from \cite{luo2021score} in the following three aspects: 1) we propose a patch-based gradient ascent method to update points in a gradient field; 2) we search the temporally corresponding patches in the gradient field of adjacent point cloud frames leveraging on rigid motion in classical
mechanics; and 3) we exploit the temporal correspondence to construct the temporal gradient field for denoising dynamic point clouds.

\section{Gradient Field Modeling For Temporal Correspondence}
\label{sec:modeling}
In this section, we will elaborate on the proposed modeling of temporal correspondence search in dynamic point clouds. 
We start from the distribution modeling of static noisy point clouds via gradient fields, and then extend such modeling to dynamic point clouds. 
Based on the modeling, we propose temporal correspondence search via gradient fields, which is a key to dynamic point cloud denoising. 
Finally, we analyze the physical meaning of the proposed model. 


\subsection{Distribution Modeling of Static Noisy Point Clouds}
\label{subsec:static_modeling}

We begin our analysis with static noisy point clouds. We denote a point cloud as $\mX = \{\vx_i\}_{i=1}^N$, where $\vx_i = [x_i, y_i, z_i]$ is the coordinate of the $i$-th point and $N$ is the total number of points. 
We can view the point cloud $\mX$ as sampled from a 3D shape supported by the underlying 2D manifold $\gM$. 
Ideally, if the sampling process does not introduce any noise, the distribution of the noise-free point cloud $p(\vx)$ is closely connected to the underlying manifold $\gM$. 
In particular, $p(\vx)$ is a 3D Dirac delta distribution ($\delta$ distribution), \ie, 
\begin{equation}
\begin{aligned}
    & p(\vx)=\left\{ 
    \begin{aligned}
        +\infty ,\  & if \  \vx\in \gM, \\
        0 ,\  & if \  \vx \notin \gM, 
    \end{aligned} \right. \\
 & \int_{-\infty}^{+\infty}\int_{-\infty}^{+\infty}\int_{-\infty}^{+\infty} p(\vx) dx dy dz = 1. 
 \end{aligned}
\end{equation}

Unfortunately, limited by the accuracy of the acquisition device and/or the sampling algorithm, point clouds often suffer from noise. 
Assuming the noise follows some distribution $n$ (\eg, Gaussian noise), then we can model the distribution about the noisy point cloud as the convolution between the original distribution $p$ and the noise model $n$, expressed as $q(\vx) = p(\vx) * n$, where $*$ is the convolution operation. 
Without the loss of generality, we assume that $n$ has a unique mode at $0$. 
Then $q(\vx)$ reaches the maximum on the manifold\footnote{We will show by experiments that the proposed model is still effective if the assumption does not hold (see  Section \ref{sec:experiments}).}, \ie, the mode of $q(\vx)$ is the underlying surface. 

Based on the above analysis, the process of denoising a point cloud $\mX = \{\vx_i\}_{i=1}^N$ naturally amounts to moving noisy points towards the mode. 
This can be formulated as:
\begin{equation}
 \max \sum_{i=1}^N \log q(\vx_i).
 \label{eq:static_formulation}
\end{equation}

However, the distribution $q(\vx_i)$ is unknown when denoising the point cloud. 
Instead, we estimate the gradient field to maximize $\sum_{i=1}^N \log q(\vx_i)$. The gradient is the first-order derivative of the log-probability function:
\begin{equation}
    \vg(\vx) = \nabla_{\vx} \log q(\vx).
    \label{eq:gradient}
\end{equation}
The gradient can be calculated at every position in the space, leading to a gradient field.
As shown in Fig.~\ref{fig:teaser}, the gradient field is a vector field, where the gradient at each location in the gradient field points to the underlying clean surface. 

Based on the gradient field, we optimize Eq.~\ref{eq:static_formulation} via gradient ascent. In each iteration, $\vx$ is updated by:
\begin{equation}
  \vx^{(h+1)} = \vx^{(h)} + \alpha^{(h)} \vg(\vx^{(h)}),   
  \label{eq:static_gradient_ascent}
\end{equation}
where $\alpha$ is the step size and $h$ indicates the index of iterations. 


\subsection{Gradient Modeling of Dynamic Noisy Point Clouds}
\label{subsec:temporal_modeling}

Further, we extend the distribution modeling in Section~\ref{subsec:static_modeling} to dynamic point clouds. 
We view a dynamic point cloud sequence with $T$ frames $\gQ=\{\mX_t\}_{t=1}^{T}$ as sampled from a 3D dynamic object/scene supported by the underlying time-varying  manifold $\gM_t$. 
If the dynamic point cloud sequence is not corrupted by noise, the points will lie exactly on the underlying dynamic manifold $\gM_t$, which follows the distribution $p_t(\vx)$. 
$p_t(\vx)$ is a time-varying 3D Dirac delta distribution: 
\begin{equation}
\begin{aligned}
    & p_t(\vx)=\left\{ 
    \begin{aligned}
        +\infty ,\  & if \  \vx\in \gM_t, \\
        0 ,\  & if \  \vx \notin \gM_t, 
    \end{aligned} \right. \\
 & \sum_{t=1}^{T}\int_{-\infty}^{+\infty}\int_{-\infty}^{+\infty}\int_{-\infty}^{+\infty} p_t(\vx) dx dy dz = 1. 
 \end{aligned}    
\end{equation} 


In order to simplify the analysis, we assume that objects will not be removed from the scene and no new objects will appear in the scene in two adjacent frames, which generally holds at sufficient sampling rates. 
Then a patch $\mP_t$ from $\mX_t$ corresponds to a patch $\mP_{t-1}$ in $\mX_{t-1}$, both of which represent the same surface in the 3D scene. 
The corresponding patches may be associated with a transformation $\gT(\cdot) = \gR(\cdot) + \gD(\cdot)$ that describes the rotation and translation of patch $\mP_t$, \ie, $\mP_{t-1} = \gT(\mP_t)$. 
Next, we will analyze the relationship between the gradient fields of $\mP_t$ and $\mP_{t-1}$.   


When suffering from noise $n$, points in dynamic point clouds follow the distribution $q_t(\vx) = p_t(\vx) * n$. 
As in Section~\ref{subsec:static_modeling}, $q_t(\vx)$ reaches the maximum on the manifold. 
Then we also model dynamic noisy point clouds with gradient fields. The gradient field of the $t$-th frame is $\nabla_{\vx} \log q_t(\vx)$.

Assuming the temporal correspondence has been searched, we propose to employ the gradient fields of adjacent frames to \textit{refine} the gradient field in the current frame, which promotes the consistency in the temporal domain for dynamic point cloud denoising. This will be discussed in detail next. 


\subsection{Correspondence Search With Gradient Field Modeling}
\label{subsec:correspondence_search}

Given a patch $\mP_t$ in the $t$-th frame, we assume its corresponding patch in the $(t-1)$-th frame is $\mP_{t-1}$. 
As discussed in Section~\ref{subsec:temporal_modeling}, $\mP_{t-1} = \gT(\mP_t)$, where $\gT$ is an affine transformation. 
Taking the previous adjacent frame as an example, our goal of the temporal correspondence search for $\mP_t$ is to find $\mP_{t-1}$ by transforming $\mP_t$ in the gradient field.    

In particular, the density of point cloud $\mX_{t-1}$ follows $q_{t-1}(\vx)$, which reaches the mode on the underlying manifold. 
Hence, we formulate the problem of searching a temporally corresponding patch pair $\mP_{t-1}$ and $\mP_{t}$ as  
\begin{equation}
    \max_{\gT(\cdot)} \sum_{\vx\in\gT(\mP_t)} q_{t-1}(\vx). 
    \label{eq:correspondence_formulation}
\end{equation}

Since $q_{t-1}(\vx)$ is difficult to estimate, similar to Section~\ref{subsec:static_modeling}, we resort to the gradient ascent via the gradient field to optimize Eq.~\ref{eq:correspondence_formulation}. 
What differs is that, we take $\mP_t$ as a whole during the gradient ascent. 
Specifically, the average of the gradients of each point in the patch is deployed as the gradient of the patch for the translation. 
Regarding the rotation, inspired by the concept of moment in classical mechanics, we let $\mP_t$ rotate as a rigid body. The force exerted on the points by the field provides the moment to rotate the entire patch. 
We will elaborate on the details in Section~\ref{sec:corr}. 

In each iteration, the location of points in patch $\mP_t$ is updated by:
\begin{equation}
    \begin{aligned}
        \vx^{(h+1)}  & =\gT^{(h)}(\vx^{(h)}) \\
        & =\gR^{(h)}(\vx^{(h)} - \bar\vx^{(h)}) + \gD^{(h)}, \vx^{(h)} \in \mP_t^{(h)},
    \end{aligned}
    \label{eq:patch_gradient_ascent}
\end{equation}
where $\gR^{(h)}$ denotes the rotation matrix relative to the patch center $\bar\vx^{(h)}$ in the $h$-th iteration, and $\gD^{(h)}$ denotes the translation. 
$\vx^{(h)}$ is iteratively updated until convergence, leading to the solution of $\gT$ in Eq.~\ref{eq:correspondence_formulation}:
\begin{equation}
\label{eq:entire_transformation}
    \gT^*(\cdot) = \gT^{(H)}(\gT^{(H-1)}(...\gT^{(1)}(\cdot)...)),
\end{equation}
where $H$ is the total number of iterations.

\begin{figure}[t]
\begin{center}
    \includegraphics[width=\columnwidth]{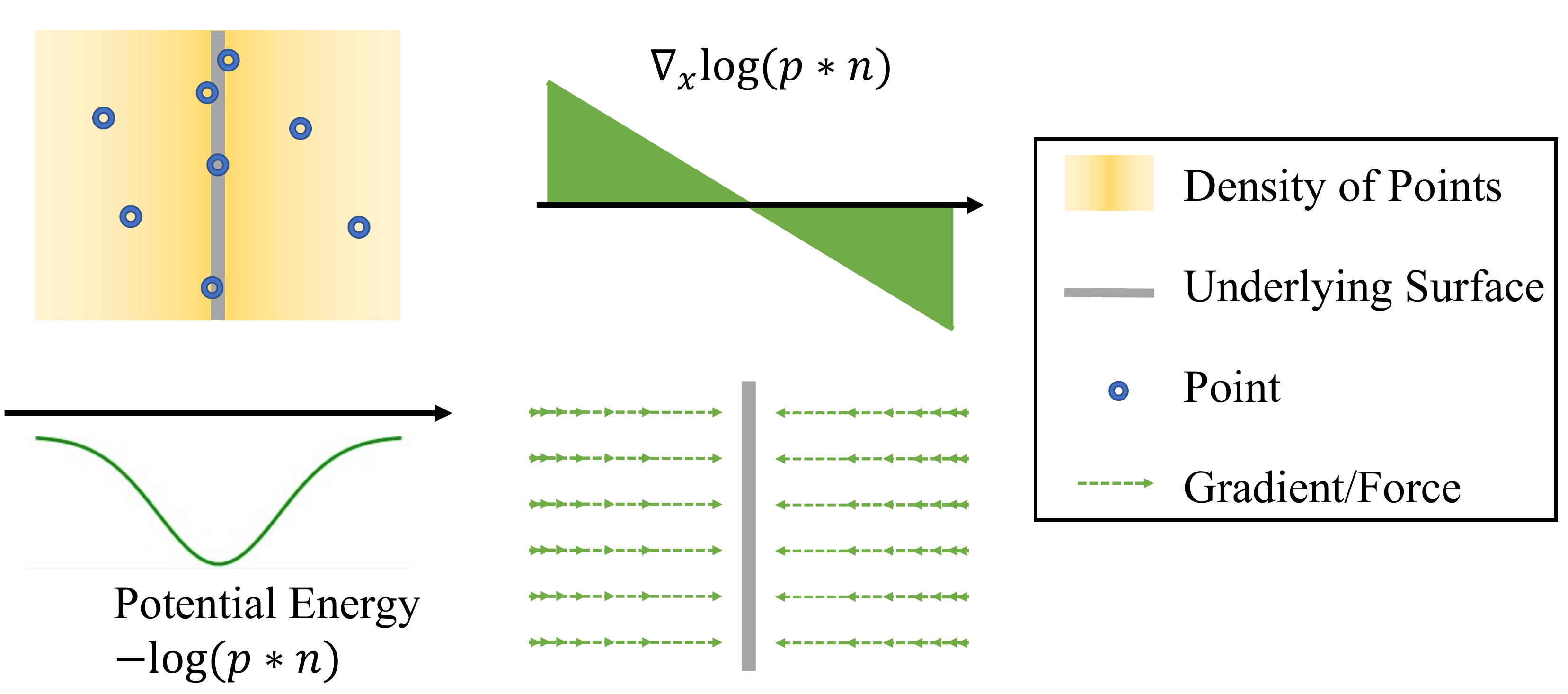}
\end{center}
\vspace{-0.1in}
  \caption{Illustration of the gradient field theory. } 
\label{fig:theory}
\end{figure}

\subsection{Analysis of Physical Meaning}
\label{subsec:phy_meaning}

Further, we provide physical insights of the aforementioned denoising process and the temporal correspondence search from the perspective of classical mechanics. 

In classical mechanics, objects in the field all have \textit{potential energy} and move under the action of \textit{force}, \eg, objects in the gravity field will fall onto the ground. 
The force on a particle in the field is equal to the negative gradient of its potential energy. 
As illustrated in Fig.~\ref{fig:theory}, we interpret gradient fields by analogy with the concepts of \textit{field}, \textit{potential energy} and \textit{force} in classical mechanics. 
In particular, we define the gradient of a point in a gradient field as a \textit{gradient force}. 
According to the definition of the gradient field in Eq.~\ref{eq:gradient}, we define a \textit{gradient potential energy} for point $\vx$ as:
\begin{equation}
    \gE(\vx) = -\log q(\vx).
    \label{eq:gradient_potential_energy}
\end{equation}

Accordingly, the denoising formulation in Eq.~\ref{eq:static_formulation} can be cast as minimizing the total gradient potential energy of the point cloud: 
\begin{equation}
    \min \sum_{i=1}^N \gE(\vx).
    \label{eq:static_formulation2}
\end{equation}

Then the gradient ascent process in Eq.~\ref{eq:static_gradient_ascent} could be treated as moving point $\vx$ with the action of the {gradient force}. In each iteration, $\vx$ moves in the direction of the gradient force, in which the moving pace is proportional to the gradient force. 

Similarly, we cast the formulation of the temporal correspondence search in Eq.~\ref{eq:correspondence_formulation} as 
\begin{equation}
    \min_{\gT(\cdot)} \sum_{\vx\in\gT(\mP_t)} \gE_{t-1}(\vx), 
    \label{eq:correspondence_formulation_2}
\end{equation}
which minimizes the \textit{gradient potential energy}. Then the patch gradient ascent process in Eq.~\ref{eq:patch_gradient_ascent} could be seen as translating and rotating the patch with the action of the gradient force. 
In particular, we employ the rigid body motion for the translation and rotation. 
The translation of the patch is proportional to the average of the gradients of points:
\begin{equation}
    \gD^{(h)} = \beta^{(h)} \frac{1}{m}\sum_{\vx^{(h)} \in \mP_t^{(h)}} \vg_{t-1}(\vx^{(h)}), 
\label{eq:rigid_body_translation}
\end{equation}
where $m$ is the number of points in the patch, $\beta^{(h)}$ is the step size of translation in the $h$-th iteration.

Regarding the rotation, let us revisit the Euler's rotation equation in Classical Mechanics:
\begin{equation}
\mathbf {I} {\dot {\boldsymbol {\omega }}}+{\boldsymbol {\omega }}\times \left(\mathbf {I} {\boldsymbol {\omega }}\right)=\mathbf{M}, 
\end{equation}
where $\mM$ is the applied torques, $\mI$ is the inertia matrix, and $\boldsymbol {\omega }$ is the angular velocity about the principal axes.
We simplify the Euler's rotation equation for our scenario:
\begin{equation}
    \mI \theta^{(h)} = \gamma^{(h)} \sum_{\vx^{(h)} \in \mP_t^{(h)}} (\vx^{(h)} - \bar\vx^{(h)}) \times \vg_{t-1}(\vx^{(h)}),
\end{equation}
where $\mI$ is the inertia matrix of the patch, $\theta^{(h)}$ is the rotation angle, $\gamma^{(h)}$ is the factor of rotation, and $\bar\vx^{(h)}$ is the patch center. 
The angular velocity $\boldsymbol{\omega }$ is simplified to $\theta^{(h)}$, and the second term on the left side of the Euler's rotation equation is simplified to $0$. 
$(\vx^{(h)} - \bar\vx^{(h)}) \times \vg_{t-1}(\vx^{(h)})$ could be seen as the moment that the gradient field performs on $\vx^{(h)}$.
Then the rotation matrix $\gR^{(h)}$ is given by:
\begin{equation}
\begin{aligned}
    \gR^{(h)} & = \bar\vx^{(h)} + \gR_\theta \left (\theta^{(h)}\right ) \\
    & = \bar\vx^{(h)} + \gR_\theta \left(\frac{\gamma^{(h)}}{\mI}  \sum_{\vx^{(h)} \in \mP_t^{(h)}} (\vx^{(h)} - \bar\vx^{(h)}) \times \vg_{t-1}(\vx^{(h)}) \right ),
\end{aligned}
\label{eq:rigid_body_rotation}
\end{equation}
where $\gR_\theta(\cdot)$ translates a rotation angle to a rotation matrix.
The rotation is relative to the patch center $\bar\vx^{(h)}$.   
In each iteration, the patch translates and rotates to a position with smaller \textit{gradient potential energy}, and finally falls onto the manifold, where the {gradient potential energy} is the smallest.


\section{The Proposed Algorithm}
\label{sec:method}
\subsection{Overview}
\label{sec:overview}


\begin{table}[t]
    \centering
    \begin{tabular}{c|c}
        \toprule 
        Notation & Description \\
        \midrule 
        $p(\cdot)$ & distribution of the clean point cloud \\
        $q(\cdot)$ & distribution of the noisy point cloud \\
        $\vg(\cdot)$ & the ground truth gradient \\
        $\gG(\cdot)$ & the initially estimated gradient \\
        $\bar\gG_{t}(\cdot)$ & the finally estimated temporal gradient \\
        $\gE(\cdot)$ & gradient potential energy \\
        $\gT(\cdot)$ & transformation between patches in adjacent frames \\
        $\gR(\cdot)$ & rotation of patches across adjacent frames \\ 
        $\gD$ & translation of patches across adjacent frames \\
        $\mI$ & inertia matrix of a patch \\
        $\gR_\theta$ & a function that transfers the angle axis to a rotation matrix \\
        $\gQ$  &  the noisy dynamic point cloud sequence \\
        $\mX_t$ & the $t$-th frame in $\gQ$ \\
        $\vx_i$ & the $i$-th point in $\mX_t$ \\
        $\mY_t$ & the $t$-th frame in the clean point cloud \\
        $\vy_i$ & the $i$-th point in $\mY_t$ \\
        $\mP_t$ & a noisy patch in frame $t$ \\
        $\gH$ & feature extraction unit \\
        $\gM$ & initial gradient estimation unit \\
        
        \bottomrule 
    \end{tabular}
    \caption{Key notations in this article.}
    \label{tab:my_label}
\end{table}

\begin{figure*}[t]
\begin{center}
    \includegraphics[width=\textwidth]{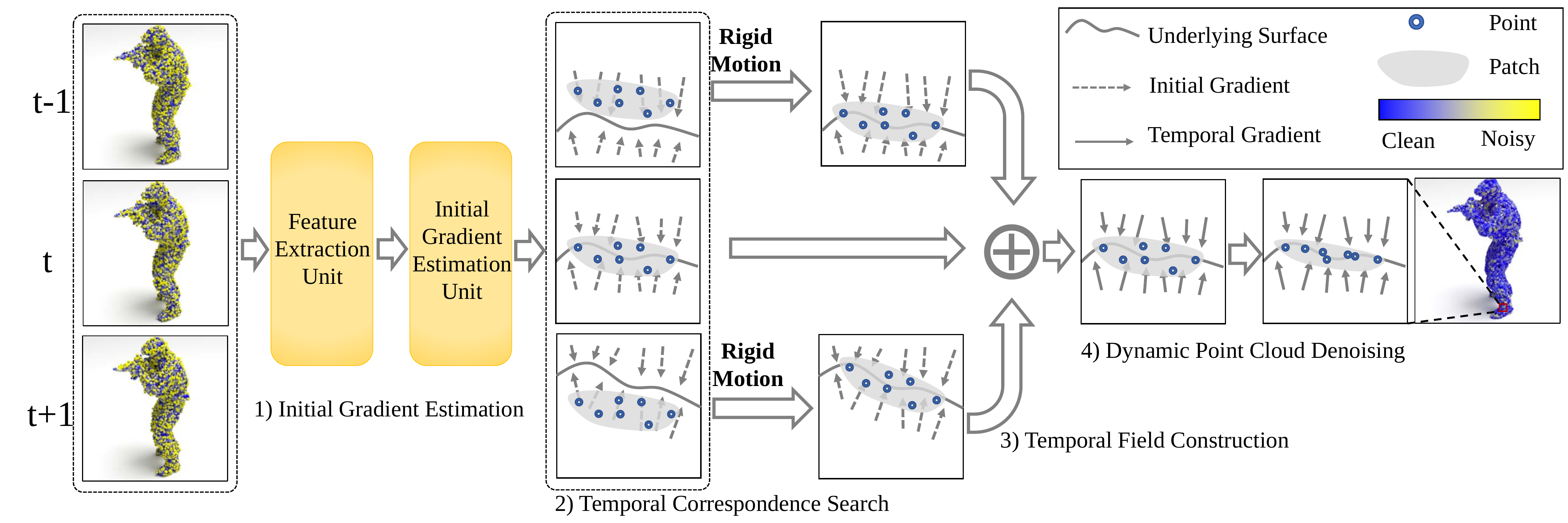}
\end{center}
\vspace{-0.1in}
  \caption{The overall framework of the proposed dynamic point cloud denoising algorithm, where the illustrated patches are the same patch from the $t$-th frame. The proposed algorithm mainly consists of four steps: 1) estimate the initial gradient field (Section~\ref{subsec:init_grad}; 2) search temporally corresponding patches via rigid motion in a classical mechanics manner (Section~\ref{sec:corr}); 3) construct the temporal gradient field by averaging the inversely transformed gradient field from corresponding patches in adjacent frames (Section~\ref{sec:field}); 4) denoise the dynamic point cloud via gradient ascent, based on the constructed temporal gradient field (Section~\ref{sec:denoising}).}
\label{fig:framework}
\end{figure*}

Given a {\it noisy} dynamic point cloud sequence $\gQ=\{\mX_1,\mX_2,...,\mX_T\}$ that consists of $T$ frames of point clouds, in which $\mX_t=[\vx_{1}^{(t)}, \vx_{2}^{(t)}, ... , \vx_{N}^{(t)}]^\top\in \mathbb{R}^{N\times3}$ corresponds to the $t$-th frame, we aim to restore the \textit{noise-free} dynamic point cloud sequence $\hat{\gQ}$.


As demonstrated in Fig.~\ref{fig:framework}, the proposed method mainly consists of four modules:
\begin{enumerate}
\setlength{\itemsep}{0pt}
\setlength{\parskip}{0pt}
    \item {\bf Initial gradient field estimation.} We first deploy a feature extraction unit to extract point-wise features of each frame. Then, we estimate the initial gradient field of each frame from point-wise features, which will be utilized in the subsequent temporal correspondence search. This module is trained with each noisy point cloud frame.
    \item \textbf{Temporal correspondence search.} We search temporally corresponding patches among adjacent frames by casting transformations of patches as rigid motions of classical mechanics in the gradient field until reaching a balanced state, \ie, when the overall gradient approaches 0;
    \item \textbf{Temporal gradient field construction.} We construct the temporal gradient field from the searched temporally corresponding patches via inverse transformations;
    \item \textbf{Dynamic point cloud denoising.} We perform gradient ascent based on the temporal gradient field for dynamic point cloud denoising.
\end{enumerate}

We elaborate on the four modules in order as follows. 

\subsection{Initial Gradient Field Estimation}
\label{subsec:init_grad}

We first estimate the initial gradient field for each frame in a dynamic point cloud sequence\footnote{In this subsection, we consider only each frame of static point cloud. For brevity, we have left out the timestamp on notations.}. We train a gradient estimation network that consists of a feature extraction unit and a gradient estimation unit.

\paragraph{Feature extraction unit} Given a noisy point cloud $\mX =\{\vx_i\}^N_{i=1}$, we first learn the context features in each local neighborhood around each point $\vx_i$. 
Particularly, we adopt the DGCNN \cite{wang2019dynamic} to build a stack of densely connected edge convolution layers for the context feature extraction, where the DGCNN is a graph convolution network commonly used in previous denoising models \cite{luo2021score,luo2020differentiable}. 
Each densely connected edge convolution layer takes the output feature of the previous layer as input (for the first layer, the input is the noisy point cloud $\mQ$). 
Then it constructs a $k$-Nearest-Neighbor (kNN) graph dynamically over the input features, where each feature item is treated as a vertex and connected to its $k$-nearest neighbors. 
Similar with \cite{huang2017densely, liu2019densepoint}, the densely connected edge convolution layer extracts rich contextual features via the dense connection of both local and non-local features. 
This property makes the extracted features suitable for point cloud denoising tasks, as evidenced in previous works \cite{luo2021score,luo2020differentiable}. 

Specifically, the feature of $\vx_i$ learned from the feature extraction unit is 
\begin{equation}
\label{eq:fnet}
  \vh_i = \gH\left(\vx_i\right),
\end{equation}
where $\gH$ denotes the feature extraction unit.

\paragraph{Gradient estimation unit} This unit is a multi-layer perceptron (MLP) that takes some 3D coordinate $\vx \in \sR^3$ nearby $\vx_i$ as input and outputs the gradient $\gG_i(\vx)$: 
\begin{equation}
\label{eq:gnet}
  \gG_i(\vx) = \gM\left(\vx - \vx_i \parallel \vh_i\right),
\end{equation}
where $\parallel$ is a concatenation operator, and $\gM$ denotes the gradient estimation unit. $\gG_i(\vx)$ is supervised by the ground truth gradient $\vg(\vx)$ defined as follows:
\begin{equation}
\label{eq:gtgradient}
    \vg(\vx) = \operatorname{NN}(\vx, \mY) - \vx, 
\end{equation}
where $\mY = \{ \vy_i \}_{i=1}^N$ is the ground truth noise-free point cloud, and $\operatorname{NN}(\vx, \mY)$ returns the point nearest to $\vx$ in $\mY$.

The optimization objective of the gradient field estimation is defined as:
\begin{equation}
  \gL^{(i)} = \E_{\vx \sim \gN(\vx_i)} \left[ \left\| \vg(\vx) - \gG_{i}(\vx)  \right\|_2^2 \right],
\end{equation}
where $\gN(\vx_i)$ is a distribution concentrated in the neighborhood of $\vx_i$ in the $\sR^3$ space, and $\E$ is the mathematical expectation. 
In practice, we generate a set $\mU$ that contains several coordinates sampled nearby $\vx_i$ and calculate the average difference of $\vg(\vx)$ and $\gG_{i}(\vx)$: 
\begin{equation}
  \gL^{(i)} = \frac{1}{\left| \mU \right|} \sum_{\vx \in \mU }\left[ \left\| \vg(\vx) - \gG_{i}(\vx)  \right\|_2^2 \right].
\label{eq:single_loss}
\end{equation}

The whole training loss is formulated as the average loss of each point:
\begin{equation}
  \gL = \frac{1}{N}\sum_{i=1}^{N}\gL^{(i)}.
\label{eq:loss}
\end{equation}

\begin{algorithm}[t]
  \caption{Training of Gradient Field Learning}  
  \label{alg:training}
  \SetKwInOut{Input}{Input}\SetKwInOut{Initialize}{Initialize}\SetKwInOut{Output}{Output}
    \Input{The noisy static point cloud $\mX=\{\vx_i\}_{i=1}^N$ and ground truth point cloud $\mY=\{\vy_i\}_{i=1}^N$}
    \Initialize{The weights of context feature extraction unit $\gH$ and the gradient field estimation unit $\gM$}
    \Repeat{convergence}{
            Learn the point-wise feature $\vh_i = \gH\left(\vx_i\right)$ \\
            \For{each $\vx_i$} {
                Generate a set $\mU$ that contains coordinates sampled nearby $\vx_i$ \\
                \For{each $\vx \in \mU$}{ 
                    Predict the gradient of $\vx$: $\gG_i(\vx) = \gM\left(\vx - \vx_i \parallel \vh_i\right)$ \\
                    Compute the ground truth gradient: $\vg(\vx) = \operatorname{NN}(\vx, \mY) - \vx$ \\
                }
                Compute the loss $\gL^{(i)}$ in Eq.~\ref{eq:single_loss} \\
            }
            Compute the total loss in Eq.~\ref{eq:ensemble} \\
            Back-propagate and update weights of $\gH$ and $\gM$
        }
        \Output{$\gH$, $\gM$ with trained weights}
\end{algorithm} 

The initial gradient field $\gG(\vx)$ for each point $\vx\in\sR^{3}$ is then computed as
\begin{equation}
\label{eq:ensemble}
  \gG(\vx) = \frac{1}{k} \sum_{\vx_i \in k\text{NN}(\vx)} \gG_{i}(\vx),
\end{equation}
where $k\text{NN}(\vx)$ is the $k$-nearest neighborhood of $\vx$. 
The training algorithm of the initial gradient field estimation is summarized in Algorithm~\ref{alg:training}.

\begin{figure}
\begin{center}
    \includegraphics[width=\columnwidth]{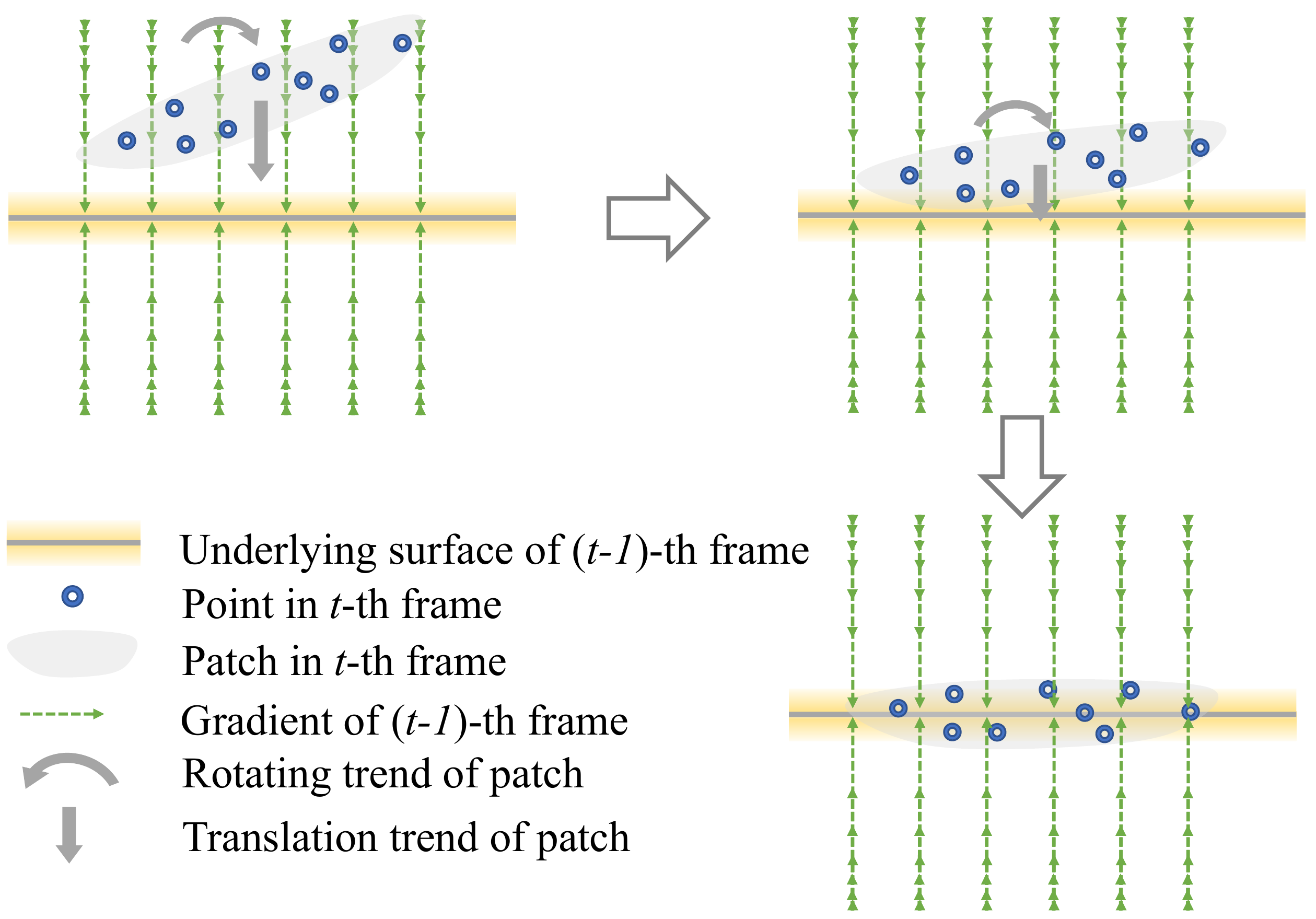}
\end{center}
\vspace{-0.1in}
  \caption{Illustration of the temporal correspondence search in the previous frame. The denser the arrow is, the larger the gradient is. Each patch is treated as a rigid body and the gradient field is treated as a force field. The target patch moves in the gradient field of an adjacent frame just like a rigid object rotating and translating under the action of the force field, until it reaches a balanced state. The balanced patch is the searched temporal correspondence.}
\label{fig:correnpondence}
\end{figure}

\subsection{Temporal Correspondence Search}
\label{sec:corr}

As discussed in Sec~\ref{subsec:phy_meaning}, in classical mechanics, the negative gradient of the potential energy of a rigid body represents the force that the rigid body receives along the direction of the gradient, which inspires us to simulate the motion of objects in a gradient field to search for the temporal correspondence, as illustrated in Fig.~\ref{fig:correnpondence}. 
In particular, we let the patch move in the gradient field of adjacent frames in order to match the local structures in consecutive frames, where the moved patch serves as \textit{the corresponding patch}. In the following, we take the temporal correspondence search between the current $t$-th frame and the previous one as an instance.

Based on the learned gradient field $\gG(\vx)$ described in Section~\ref{subsec:init_grad}, we denote the gradient field in the $t$-th frame as $\gG_{t}(\vx)$, $t=1,...,T$.
To consider the temporal correspondence search for the current noisy point cloud $\mX_{t}$, we first divide $\mX_{t}$ into $M$ overlapped patches, whose patch centers are selected from a subset of points $\{\vc_i\}_{i=1}^{M}$ in $\mX_{t}$ by the farthest point sampling (FPS) algorithm \cite{fps}.
Next, for each center point $\vc_i$, we construct a patch by identifying the $(m-1)$-nearest neighbors of $\vc_i$, leading to $M$ patches.

Given a noisy patch $\mP_t=\{\vx_i\}_{i=1}^{m}$ in frame $t$, we aim to search its corresponding patch in the previous frame. 
As discussed in Section~\ref{subsec:correspondence_search}, we perform patch gradient ascent to let the patch move onto the position of its correspondence, as demonstrated in Fig.~\ref{fig:correnpondence}. 
As discussed, we treat the iterative patch gradient ascent process as a series of rigid motions under the action of the gradient field. 
In particular, the rigid motions of moving a patch in the gradient field include \textit{rotation} and \textit{translation}. 
Then we iteratively update the position of patch $\mP_t$ as a whole via Eq.~\ref{eq:patch_gradient_ascent}. 
The details of learning patch rotation and translation are elaborated as follows.

\begin{algorithm}[t]
  \caption{The Proposed Dynamic Point Cloud Denoising}  
  \label{alg:denoising}
  \SetKwInOut{Input}{Input}\SetKwInOut{Output}{Output}
    \Input{The noisy point cloud sequence $\gQ=\{\mX_t\}_{t=1}^T$}
    \For{$\mX_t$ in $\gQ$}{
        Calculate the initial gradient field $\gG_t(\vx)$ with Eq.~\ref{eq:ensemble} \\
    }
    \For{$\mX_t$ in $\gQ$}{
            $\gS \leftarrow \emptyset$ \\
            Select $M$ patch centers $\{\vc_i\}_{i=1}^{M}$ with the FPS algorithm \\
            Construct $M$ patches $\{\mP_i\}_{i=1}^{M}$ for each $\vc_i$ with $\vc_i$'s $(m-1)$-nearest neighbors\\
            \For {each patch $\mP_i$ in the $t$-th frame} {
                \For {$tt$ in $\{t-1, t+1\}$} {
                    Search the temporal correspondence of $\mP_i$ in the $tt$-th frame with the  algorithm in Section~\ref{subsec:correspondence_search} \\
                    Calculate $\gG_{tt}'(\cdot) $ with Eq.~\ref{eq:transformed_gradient} \\
                }
                $\bar\gG_t(\cdot) = \frac{1}{3}(\gG_{t-1}'(\cdot)+\gG_t(\cdot)+\gG_{t+1}'(\cdot))$ \\
                $\{\vx_j^{(0)}\}_{j=1}^{m} \leftarrow \mP_i$ \\
                Update $\vx_j$ with gradient ascent as in Eq.~\ref{eq:denoise} \\
                $\gS \leftarrow \gS \bigcup \{\vx_j^{(H')}\}_{j=1}^{m}$ \\
            }
            Downsample $\gS$ with FPS to acquire the denoised point cloud $\bar\mX_t$ \\
    }
    \Output{The denoised point cloud sequence $\bar\gQ=\{\bar\mX_t\}_{t=1}^T$}
\end{algorithm} 

\paragraph{Patch Rotation}
In each iteration, the rotation angle is calculated by:
\begin{equation}
    \mI \theta^{(h)} = \gamma^{(h)} \sum_{\vx^{(h)} \in \mP_t^{(h)}} (\vx^{(h)} - \bar\vx^{(h)}) \times \gG_{t-1}(\vx^{(h)}),
\label{eq:rigid_body_rotation_estimated}
\end{equation}
where the rotation is relative to the patch center $\bar\vx$. 
This distinguishes from Eq.~\ref{eq:rigid_body_rotation} in that we employ the estimated gradient field, because the ground truth gradient is unavailable. 

In practice, we set each point to be of unit mass. Thus, the inertia matrix is given by
\begin{equation}
\begin{aligned}
\mathbf{I}=&\left[\begin{array}{ccc}
I_{x x} & I_{x y} & I_{x z} \\
I_{y x} & I_{y y} & I_{y z} \\
I_{z x} & I_{z y} & I_{z z}
\end{array}\right] \\
=&\left[\begin{array}{ccc}
\sum_{j=1}^{m} \left(y_{j}^{2}+z_{j}^{2}\right) & -\sum_{j=1}^{m}  x_{j} y_{j} & -\sum_{j=1}^{m}  x_{j} z_{j} \\
-\sum_{j=1}^{m}  x_{j} y_{j} & \sum_{j=1}^{m} \left(x_{j}^{2}+z_{j}^{2}\right) & -\sum_{j=1}^{m}  y_{j} z_{j} \\
-\sum_{j=1}^{m}  x_{j} z_{j} & -\sum_{j=1}^{m}  y_{j} z_{j} & \sum_{j=1}^{m} \left(x_{j}^{2}+y_{j}^{2}\right)
\end{array}\right], 
\end{aligned}
\end{equation}
where $x_j, y_j, z_j$ are the x-, y-, z-coordinate relative to the patch center $\bar\vx$. 
Then the rotation angle relative to the patch center is
\begin{equation}
    \theta^{(h)} = \gamma^{(h)}\mI^{-1} \sum_{\vx^{(h)} \in \mP_t^{(h)}} (\vx^{(h)} - \bar\vx^{(h)}) \times \gG_{t-1}(\vx^{(h)}).
\end{equation}

Hence, the position of the rotated patch is:
\begin{equation}
    \gR^{(h)}(\mP_t^{(h)}) = \bar\vx^{(h)} + \gR_\theta(\theta^{(h)})(\vx^{(h)} - \bar\vx^{(h)}),
    \label{eq:final_rotaion}
\end{equation}
where $\gR_\theta$ transfers the angle axis to the rotation matrix. Note that, the rotation is relative to the patch center $\bar\vx^{(h)}$, so we add the absolute coordinate of the patch center in Eq.~\ref{eq:final_rotaion}.

\paragraph{Patch Translation}

The patch translation is calculated by
\begin{equation}
    \gD^{(h)} = \beta^{(h)} \frac{1}{m}\sum_{\vx^{(h)} \in \mP_t^{(h)}} \gG_{t-1}(\vx^{(h)}). 
\label{eq:rigid_body_translation_estimated}
\end{equation}
Here we replace $\vg$ in Eq.~\ref{eq:rigid_body_translation} with $\gG$, because we only have access to the estimated gradient. 

With the computed rotation and translation, the transformation of the patch in each iteration is given by
\begin{equation}
\begin{aligned}
    \mP_t^{(h+1)} &= \gR^{(h)}(\mP_t^{(h)}) + \gD^{(h)}, h = 1, 2, 3, ... \\
    \mP_t^{(0)} &= \mP_t.
\end{aligned}
\label{eq:update_patch}
\end{equation}

The transformation process iterates until the sum of gradients over the patch
reaches $0$.
Then, the transformed patch $\gT^*(\mP)$ falls onto the corresponding patch in the $(t-1)$-th frame. 

Similarly, we search the corresponding patch of $\mP_t$ in the $(t+1)$-th frame in the same way.

\begin{figure*}[t]
\begin{center}
    \includegraphics[width=\textwidth]{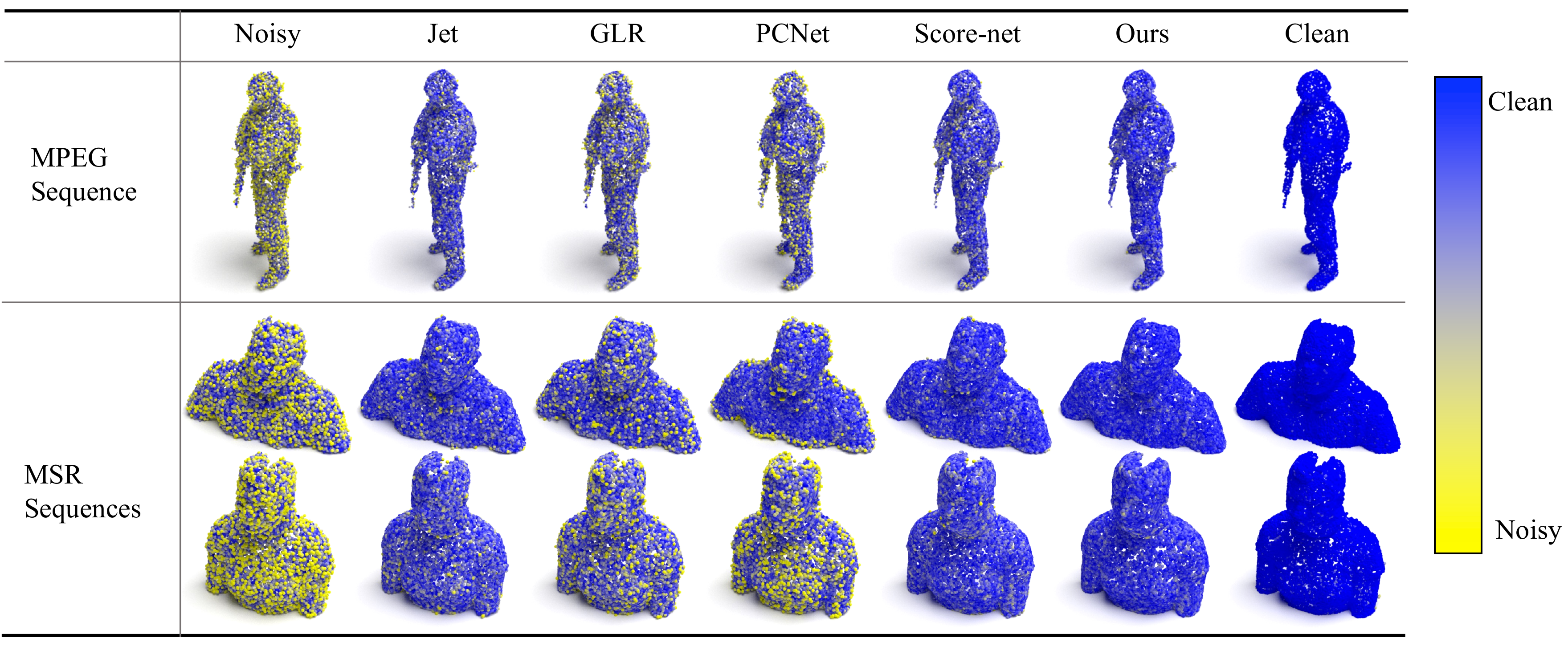}
\end{center}
\vspace{-0.1in}
  \caption{Visual comparison of the proposed method and competitive denoising methods under Gaussian noise. Points with yellower color are farther away from the ground truth surface.}
\label{fig:visualize}
\end{figure*}

\subsection{Temporal Gradient Field Construction}
\label{sec:field}

In order to construct the temporal gradient field, we first transform the gradient field of adjacent frames to the target frame, where we consider each patch separately. 
For a patch $\mP_t$ in the $t$-th frame, we denote its correspondence in adjacent frames as $\gT_{t-1}(\mP_t)$ and $\gT_{t+1}(\mP_t)$. 
We transform the gradient field of $\gT_{t-1}(\mP)$ and $\gT_{t+1}(\mP)$ to the coordinate system of the $t$-th frame. 
Specifically, taking $\gT_{t-1}(\mP)$ as an example, the transformed gradient field is:
\begin{equation}
    \gG_{t-1}'=\gR_{t-1}^{-1}\gG_{t-1}(\gT_{t-1}(\vx)),
    \label{eq:transformed_gradient}
\end{equation}
where $\gR_{t-1}$ is the rotation corresponding to $\gT_{t-1}$. 
$\gG_{t-1}'$ and $\gG_{t-1}$ have different directions but with the same magnitude, so the inverse transformation is only reflected in the rotation.

Considering the target frame and its previous and subsequent frames, the temporal gradient field of patch $\mP_t$ is the average gradients over the temporally corresponding patches $\{\mP_{t-1},\mP_{t},\mP_{t+1}\}$ in the three consecutive frames,
\begin{equation}
\label{eq:final-field}
\bar\gG_{t}(\vx) = \frac{1}{3} \bigg(\gG_{t}(\vx) + \gG_{t-1}'(\vx)  + \gG_{t+1}'(\vx)\bigg).
\end{equation}

\subsection{Dynamic Point Cloud Denoising}
\label{sec:denoising}

Based on the estimated temporal gradient field, we deploy the gradient ascent algorithm to update the position of each point: 
\begin{equation}
\label{eq:denoise}
\begin{split}
    \vx_j^{(h)} & = \vx_j^{(h-1)} + \alpha^{(h)} \bar\gG_{t}(\vx_j^{(h-1)}), \ h = 1,\ldots,H', \\
    \vx_j^{(0)} & = \vx_j, \ \vx_j \in \mP_t ,
\end{split}
\end{equation}
where $\alpha^{(h)}$ is the step size at the $h$-th iteration, and $H'$ is the total number of iterations. 

In particular, we denoise each patch separately. 
Specifically, we first initialize an empty set $\gS$. 
When we finish denoising a patch, we add the denoised points to the set $\gS$. Note that, the patches are overlapped with each other, so the the final point set in $\gS$ is larger than $\mX_t$. 
Hence, we sample $\gS$ to the same cardinality of $\mX_t$ via FPS, leading to the denoised point cloud $\bar\mX_t$. 
We perform denoising of each frame like this, and finally acquire the denoised point cloud sequence $\bar\gQ=\{\bar\mX_t\}_{t=1}^T$.  
We summarize the proposed dynamic point cloud denoising algorithm in Algorithm~\ref{alg:denoising}.

\section{Experiments}
\label{sec:experiments}
\begin{table*}[t]
\begin{center}

\begin{tabular}{cl|ccc|ccc|ccc|ccc}
\toprule
\multicolumn{2}{c|}{Noise}                                                & \multicolumn{3}{c|}{0.60\%}                       & \multicolumn{3}{c|}{1.00\%}                       & \multicolumn{3}{c|}{2.00\%}                       & \multicolumn{3}{c}{3.00\%}                       \\ 
\multicolumn{1}{c|}{Dataset}                & \multicolumn{1}{c|}{method} & CD             & HD             & P2M            & CD             & HD             & P2M            & CD             & HD             & P2M            & CD             & HD             & P2M            \\
\midrule
\multicolumn{1}{c|}{\multirow{10}{*}{MSR}}  & Bilateral                   & 2.277          & 0.377          & 1.289          & 2.243          & 0.409          & 1.241          & 2.258          & 0.495          & 1.290          & 6.316          & 1.166          & 4.871          \\
\multicolumn{1}{c|}{}                       & Jet                         & 0.550          & 0.056          & 0.082          & 0.756          & 0.125          & 0.176          & 2.298          & 0.483          & 1.295          & 5.558          & 1.137          & 4.180          \\
\multicolumn{1}{c|}{}                       & MRPCA                       & 0.536          & 0.081          & 0.082          & 0.635          & 0.123          & 0.122          & 2.185          & 0.396          & 1.176          & 6.042          & 0.936          & 4.527          \\
\multicolumn{1}{c|}{}                       & GLR                         & 0.547          & 0.053          & 0.084          & 0.890          & 0.167          & 0.272          & 3.391          & 0.692          & 2.310          & 8.630          & 1.547          & 7.185          \\
\multicolumn{1}{c|}{}                       & M2M                         & 0.658          & 0.087          & 0.136          & 0.791          & 0.261          & 0.193          & 4.295          & 0.686          & 3.114          & 8.808          & 1.532          & 7.361          \\
\cmidrule{2-14}
\multicolumn{1}{c|}{}                       & TotalDn                     & 0.588          & 0.050          & 0.098          & 0.983          & 0.165          & 0.324          & 4.127          & 0.686          & 2.924          & 8.750          & 1.549          & 7.282          \\
\multicolumn{1}{c|}{}                       & DMR                         & 1.016          & 1.447          & 0.324          & 1.037          & 1.420          & 0.338          & 1.393          & 1.366          & 0.620          & 3.707          & 1.418          & 2.307          \\
\multicolumn{1}{c|}{}                       & PCNet                       & 0.794          & 0.137          & 0.214          & 1.094          & 0.204          & 0.415          & 2.590          & 0.929          & 1.701          & 7.592          & 1.980          & 6.204          \\
\multicolumn{1}{c|}{}                       & Score-Net                   & 0.527          & 0.048          & 0.075          & 0.645          & 0.120          & 0.123          & 1.473          & 0.583          & 0.679          & 3.943          & 1.414          & 2.730          \\
\cmidrule{2-14}
\multicolumn{1}{c|}{}                       & Ours                        & \textbf{0.524} & \textbf{0.044} & \textbf{0.073} & \textbf{0.633} & \textbf{0.105} & \textbf{0.119} & \textbf{1.330} & \textbf{0.575} & \textbf{0.574} & \textbf{3.425} & \textbf{1.202} & \textbf{2.281} \\
\midrule
\midrule
\multicolumn{1}{c|}{\multirow{10}{*}{MPEG}} & Bilateral                   & 0.868          & 0.051          & 0.257          & 1.067          & 0.125          & 0.379          & 2.677          & 0.525          & 1.715          & 6.442          & 1.217          & 5.199          \\
\multicolumn{1}{c|}{}                       & Jet                         & 0.623          & 0.041          & 0.103          & 0.879          & 0.113          & 0.246          & 2.494          & 0.486          & 1.540          & 5.631          & 1.143          & 4.431          \\
\multicolumn{1}{c|}{}                       & MRPCA                       & 0.637          & 0.044          & 0.112          & 0.761          & 0.085          & 0.172          & 2.228          & 0.589          & 1.291          & 5.821          & 1.243          & 4.530          \\
\multicolumn{1}{c|}{}                       & GLR                         & 0.631          & 0.049          & 0.114          & 1.046          & 0.164          & 0.377          & 3.952          & 0.665          & 2.918          & 8.522          & 1.479          & 7.297          \\
\multicolumn{1}{c|}{}                       & M2M                         & 0.801          & 0.062          & 0.209          & 1.070          & 0.160          & 0.379          & 4.410          & 0.659          & 3.344          & 8.423          & 1.467          & 7.210          \\
\cmidrule{2-14}
\multicolumn{1}{c|}{}                       & TotalDn                     & 0.666          & 0.044          & 0.122          & 0.947          & 0.155          & 0.293          & 3.775          & 0.647          & 2.701          & 8.009          & 1.467          & 6.749          \\
\multicolumn{1}{c|}{}                       & DMR                         & 1.418          & 0.398          & 0.645          & 1.541          & 0.401          & 0.745          & 2.113          & 0.529          & 1.235          & 4.667          & 1.428          & 3.595          \\
\multicolumn{1}{c|}{}                       & PCNet                       & 0.921          & 0.106          & 0.285          & 1.077          & 0.182          & 0.387          & 2.153          & 0.899          & 1.227          & 5.860          & 2.148          & 4.429          \\
\multicolumn{1}{c|}{}                       & Score-Net                   & 0.628          & 0.037          & 0.104          & 0.797          & 0.091          & 0.183          & 1.976          & 0.622          & 1.115          & 4.829          & 1.439          & 3.690          \\
\cmidrule{2-14}
\multicolumn{1}{c|}{}                       & Ours                        & \textbf{0.621} & \textbf{0.036} & \textbf{0.099} & \textbf{0.761} & \textbf{0.083} & \textbf{0.170} & \textbf{1.824} & \textbf{0.510} & \textbf{0.986} & \textbf{4.442} & \textbf{1.236}          & \textbf{3.329} \\
\bottomrule
\end{tabular}
\end{center}
\caption{Comparison of denoising Gaussian noise among competitive denoising algorithms. CD is multiplied by $10^4$, HD is multiplied by $10^2$ and P2M is multiplied by $10^4$. M2M and ours are designed for dynamic point cloud denoising.}
\label{table:quantitative}
\end{table*}

\begin{table*}[]
\begin{center}
\begin{tabular}{cl|ccc|ccc|ccc|ccc}
\toprule
\multicolumn{2}{c|}{Noise}                                               & \multicolumn{3}{c|}{0.60\%}                      & \multicolumn{3}{c|}{1.00\%}                      & \multicolumn{3}{c|}{2.00\%}                      & \multicolumn{3}{c}{3.00\%}                         \\
\multicolumn{1}{c|}{Dataset}                & \multicolumn{1}{c|}{method} & CD             & HD             & P2M            & CD             & HD             & P2M            & CD             & HD             & P2M            & CD              & HD             & P2M             \\
\midrule
\multicolumn{1}{c|}{\multirow{10}{*}{MSR}}  & Bilateral                  & 1.311          & 0.666          & 0.592          & 3.354          & 1.258          & 2.374          & 14.449         & 4.290          & 13.028         & 32.810          & 9.777          & 31.193          \\
\multicolumn{1}{c|}{}                       & Jet                        & 1.345          & 0.686          & 0.607          & 3.246          & 1.283          & 2.252          & 13.426         & 4.258          & 12.014         & 31.006          & 9.678          & 29.393          \\
\multicolumn{1}{c|}{}                       & MRPCA                      & 1.222          & 0.663          & 0.521          & 3.331          & 1.230          & 2.313          & 14.313         & 4.215          & 12.835         & 32.593          & 9.490          & 30.910          \\
\multicolumn{1}{c|}{}                       & GLR                        & 1.508          & 0.689          & 0.762          & 3.930          & 1.323          & 2.927          & 15.978         & 4.457          & 14.583         & 35.057          & 10.070         & 33.484          \\
\multicolumn{1}{c|}{}                       & M2M                        & 1.510          & 0.684          & 0.768          & 2.768          & 1.253          & 1.800          & 16.227         & 4.474          & 14.814         & 34.967          & 10.112         & 33.393          \\
\cmidrule{2-14}
\multicolumn{1}{c|}{}                       & TotalDn                    & 1.649          & 0.719          & 0.855          & 4.453          & 1.347          & 3.377          & 16.230         & 4.491          & 14.813         & 35.023          & 10.145         & 33.446          \\
\multicolumn{1}{c|}{}                       & DMR                        & 1.084          & 1.543          & 0.391          & 2.486          & 1.468          & 1.419          & 10.029         & 4.254          & 9.938          & 26.847          & \textbf{8.948} & 25.543          \\
\multicolumn{1}{c|}{}                       & PCNet                      & 1.126          & 0.587          & 0.466          & 2.727          & 1.969          & 1.887          & 13.130         & 6.423          & 11.394         & 34.949          & 11.808         & 32.842          \\
\multicolumn{1}{c|}{}                       & Score-Net                  & 1.014          & 0.650          & 0.368          & 2.314          & 1.090          & 1.465          & 11.009         & 4.347          & 9.689          & 27.471          & 9.972          & 25.929          \\
\cmidrule{2-14}
\multicolumn{1}{c|}{}                       & Ours                       & \textbf{0.970} & \textbf{0.632} & \textbf{0.353} & \textbf{2.253} & \textbf{1.056} & \textbf{1.374} & \textbf{9.798} & \textbf{4.181} & \textbf{9.233} & \textbf{24.389} & 9.131          & \textbf{22.327} \\
\midrule
\midrule
\multicolumn{1}{c|}{\multirow{10}{*}{MPEG}} & Bilateral                  & 1.828          & 0.341          & 0.936          & 3.527          & 0.941          & 2.491          & 12.094         & 3.877          & 10.817         & 24.639          & 8.986          & 23.263          \\
\multicolumn{1}{c|}{}                       & Jet                        & 1.740          & 0.360          & 0.865          & 3.420          & 0.962          & 2.376          & 11.224         & 3.886          & 9.942          & 22.999          & 8.944          & 21.619          \\
\multicolumn{1}{c|}{}                       & MRPCA                      & 1.558          & 0.320          & 0.720          & 3.228          & 0.861          & 2.196          & 11.552         & 3.628          & 10.229         & 23.833          & 8.567          & 22.395          \\
\multicolumn{1}{c|}{}                       & GLR                        & 1.873          & 0.366          & 0.995          & 4.022          & 1.002          & 2.986          & 13.447         & 4.005          & 12.198         & 26.306          & 9.070          & 24.987          \\
\multicolumn{1}{c|}{}                       & M2M                        & 1.886          & 0.358          & 1.023          & 3.027          & \textbf{0.876} & 2.021          & 13.490         & 4.022          & 12.244         & 26.119          & 9.116          & 24.806          \\
\cmidrule{2-14}
\multicolumn{1}{c|}{}                       & TotalDn                    & 1.757          & 0.389          & 0.868          & 4.055          & 0.992          & 2.978          & 13.171         & 3.935          & 11.895         & 25.849          & 9.040          & 24.503          \\
\multicolumn{1}{c|}{}                       & DMR                        & 2.140          & 0.500          & 1.179          & 3.293          & 1.000          & 1.982          & 10.702         & \textbf{3.523} & 9.530          & 23.064          & 9.215          & 20.799          \\
\multicolumn{1}{c|}{}                       & PCNet                      & 1.819          & 0.425          & 0.920          & 3.229          & 1.352          & 2.137          & 12.854         & 4.843          & 11.236         & 28.261          & 9.847          & 26.277          \\
\multicolumn{1}{c|}{}                       & Score-Net                  & 1.603          & 0.329          & 0.743          & 3.009          & 0.962          & 1.991          & 10.386         & 4.004          & 9.121          & 22.124          & 9.164          & 20.755          \\
\cmidrule{2-14}
\multicolumn{1}{c|}{}                       & Ours                       & \textbf{1.580} & \textbf{0.305} & \textbf{0.710} & \textbf{2.842} & 0.926 & \textbf{1.864} & \textbf{9.673} & 3.574          & \textbf{8.284} & \textbf{20.510} & \textbf{8.523} & \textbf{19.911} \\
\bottomrule
\end{tabular}
\end{center}
\caption{Comparison of denoising simulated LiDAR noise among competitive denoising algorithms. CD is multiplied by $10^4$, HD is multiplied by $10^2$ and P2M is multiplied by $10^4$. M2M and ours are designed for dynamic point cloud denoising.}
\label{table:scanner}
\end{table*}

\begin{figure*}[]
\begin{center}
    \includegraphics[width=\textwidth]{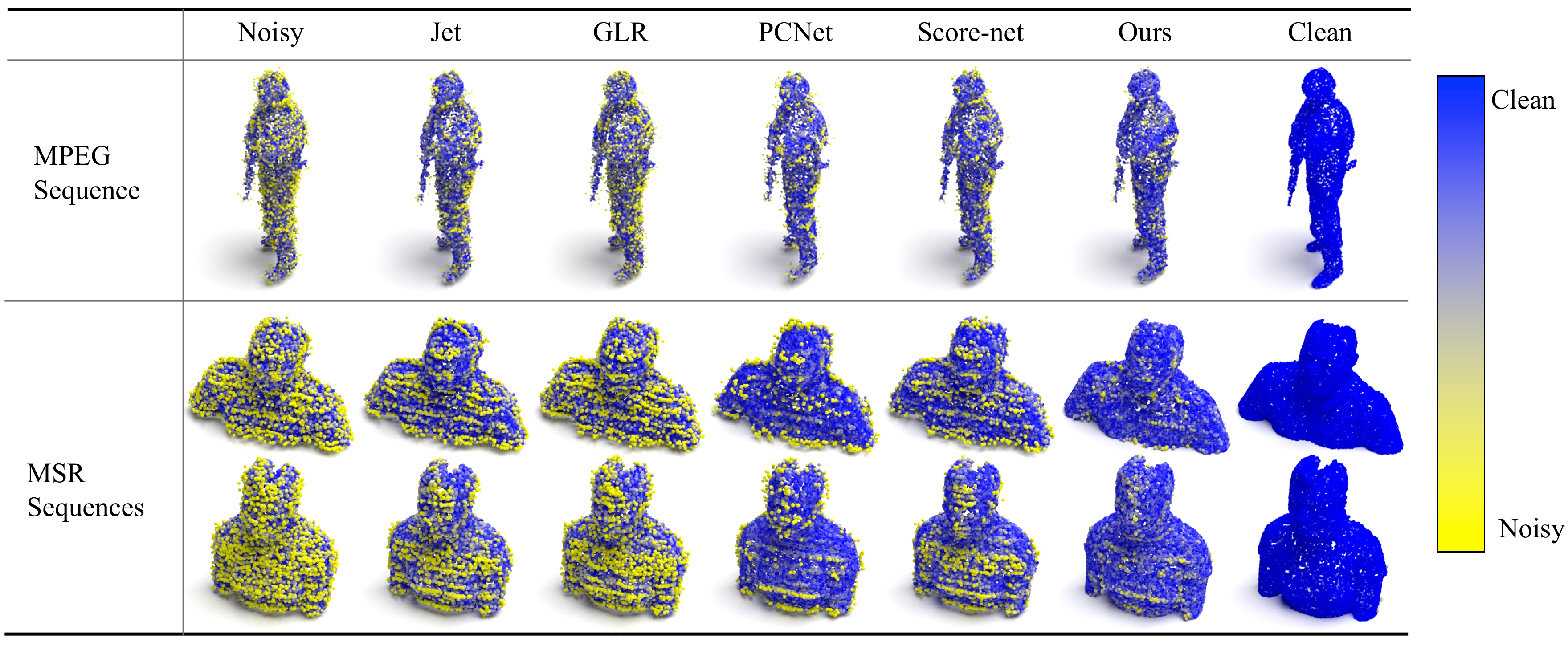}
\end{center}
\vspace{-0.1in}
  \caption{Visual comparison of the proposed method and competitive denoising methods under simulated LiDAR noise. Points with yellower color are farther away from the ground truth surface.}
\label{fig:visualize_scanner}
\end{figure*}

\subsection{Experimental Setup}

\noindent\textbf{Datasets.}
As there is no standard dataset for the dynamic point cloud denoising yet, we construct our datasets from eight dynamic point cloud sequences of two benchmarks, including four MPEG sequences from \cite{MPEG}: \textit{Longdress}, \textit{Loot}, \textit{Redandblack} and \textit{Soldier}, and five MSR sequences from \cite{Cai16}: \textit{Andrew}, \textit{David}, \textit{Phil}, \textit{Ricardo}, and \textit{Sarah}. 
In the MPEG dataset, the sequences of \textit{Longdress}, \textit{Loot}, and \textit{Redandblack} are used for training, while the sequence of \textit{Soldier} is used for testing.
In the MSR dataset, the sequences of \textit{Andrew}, \textit{David} and \textit{Phil} are used for training, while those of \textit{Ricardo} and \textit{Sarah} are used for testing.
Each point cloud frame in the sequences is downsampled to $30K$ points via the farthest point sampling algorithm. 

\noindent\textbf{Baselines.}
We compare our method with competitive denoising approaches, including five optimization-based methods: bilateral filtering \cite{digne2017bilateral}, Jet fitting \cite{cazals2005jetsfit}, MRPCA \cite{mattei2017MRPCA}, GLR \cite{zeng2019GLR} and M2M \cite{hu2021dynamic}, as well as four deep-learning-based methods: TotanDn \cite{TotalDenoising2019}, DMRDenoise (DMR) \cite{luo2020DMR}, PointCleanNet (PCNet) \cite{rakotosaona2020PCN}, and Score-Net \cite{luo2021score}.
Note that, among the competitive methods, only M2M is designed for dynamic point cloud denoising, while the others are for static point clouds and conducted frame by frame to denoise dynamic point clouds.  

\noindent\textbf{Metrics.}
We adopt three commonly used metrics in previous works to quantitatively evaluate our model: 
1) Chamfer Distance (CD) \cite{fan2017pointsetgen}, which measures the average distance from each denoised point to its nearest ground truth point; 
2) Hausdorff Distance (HD) \cite{Huttenlocher93}, which measures the farthest outlier's distance to ground truth points; and 
3) Point-to-Mesh distance (P2M) \cite{ravi2020pytorch3d}, which measures the average distance from points to the underlying clean surface. 
Note that, we normalize each denoised point cloud into the unit sphere for each method before computing the metrics.

\noindent\textbf{Implementation Details.}
The feature extractor mentioned in Section~\ref{subsec:init_grad} is composed of a four-layer stacked and densely connected dynamic graph convolutional layers \cite{wang2019dynamic}. The patch size $m$ is set to $1000$. 
In the process of model training, $k$ in Eq.~\ref{eq:ensemble} is set to $32$, the learning rate is set to $10^{-4}$ and the weight decay is $0$. 
In the process of denoising, the number of patch centers $M$ is $90$, the number of iterations $H$ and $H'$ are both set to $50$, the translation and rotation factors of the correspondence search are set as $\beta^{(h)} = 0.01 \cdot 0.95^h, h=1,...,H' $, $\gamma^{(h)} = 0.01 \cdot 0.95^h, h=1,...,H' $. 
The step size sequence in Eq.~\ref{eq:denoise} is obtained by $\alpha^{(h)} = 0.008 \cdot 0.95^h, h=1,...,H' $. 

\subsection{Quantitative Results}
We provide quantitative results of point cloud denoising on synthetic noise and simulated real-world noise, respectively.   

\subsubsection{Results on Synthetic Noise}
For the MPEG dataset, we train our model on the three sequences (\textit{Longdress, Loot, Redandblack}) perturbed by Gaussian noise with random standard deviation in the range $[0.6\%,3.0\%]$ and evaluate the denoising performance of our model on the \textit{Soldier} sequence perturbed by Gaussian noise with standard deviation in $\{0.6\%,1.0\%,2.0\%,3.0\%\}$. 
For the MSR dataset, we train our model on the three sequences (\textit{Andrew, David, Phil}) perturbed by Gaussian noise with random standard deviation in the range $[0.6\%,3.0\%]$ and evaluate the denoising performance of our model on the \textit{Ricardo} sequence and the \textit{Sarah} sequence perturbed by Gaussian noise with standard deviation in $\{0.6\%,1.0\%,2.0\%,3.0\%\}$.

As presented in Table~\ref{table:quantitative}, our model outperforms state-of-the-art optimization-based methods and deep-learning-based methods on the two datasets under all noise levels, which validates the effectiveness of our method. 
Further, we observe that we achieve larger gains at higher noise levels.
This is because the temporal correspondence searched by the proposed algorithm becomes more useful when the noise level is high.
When the noise level is relatively low, the static point cloud itself retains sufficient information for denoising; when the noise level becomes higher, each point cloud frame loses more structural information, so the information from the temporal correspondence becomes important. 

\subsubsection{Results on Simulated Real-World Noise}

Due to the lack of real-world noisy datasets of dynamic point clouds, we simulate the noise produced by LiDAR sensors for the evaluation of our method and competitive denoising methods. 

We simulate LiDAR scanner noise as follows. 
We first construct a mesh from the original point cloud with MeshLab \cite{Meshlab}. 
Then we scan the mesh with a LiDAR simulator ``Blensor" \cite{gschwandtner2011blensor} to produce a point cloud in a LiDAR-acquired style. 
In particular, to scan the mesh, we normalize the mesh into a unit sphere and put two virtual cameras in each of the six directions (top, bottom, left, right, front, and back) of the box for scanning. 
The mesh is then scanned in a Velodyne type, where the Velodyne model is set to ``hdl64e2" and the Velodyne angle resolution is set to 0.1. 
The noise level is set to $\{0.6\%, 1.0\%, 2.0\%, 3.0\%\}$. 

As shown in Table~\ref{table:scanner}, our model outperforms state-of-the-art optimization-based methods and deep-learning-based methods on the two datasets under most noise levels. 
This shows that the proposed model is still effective even if the assumption for noise in Section~\ref{subsec:static_modeling} does not hold.

\subsection{Qualitative Results}
We visualize the denoising results on synthetic noise and simulated LiDAR noise, respectively. 
Further, we provide the visualization of the temporal correspondence search results for a better understanding.

\subsubsection{Visualization of denoising results}
Fig.~\ref{fig:visualize} visualizes the denoising results of our method and competitive baselines.
The colored points illustrate the reconstruction error measured by the point-to-mesh distance.
Points with larger point-to-mesh distances (larger error) are colored brighter toward yellow, while points with smaller point-to-mesh distances (smaller error) are colored deeper toward blue. 
We see that our results are much cleaner than the baseline methods.
Notably, detailed regions such as the corners of the arms and gun are well denoised by our algorithm, while other approaches still suffer from noise and even outliers.
This validates the superiority of the proposed denoising algorithm.

Also, we visualize the denoising results on simulated LiDAR noise. 
As illustrated in Fig.~\ref{fig:visualize_scanner}, the results of our method are much cleaner than those of competitive denoising methods. 
This further validates the effectiveness of our method under simulated real-world noise.

\begin{figure}[t]
\begin{center}
    \includegraphics[width=\columnwidth]{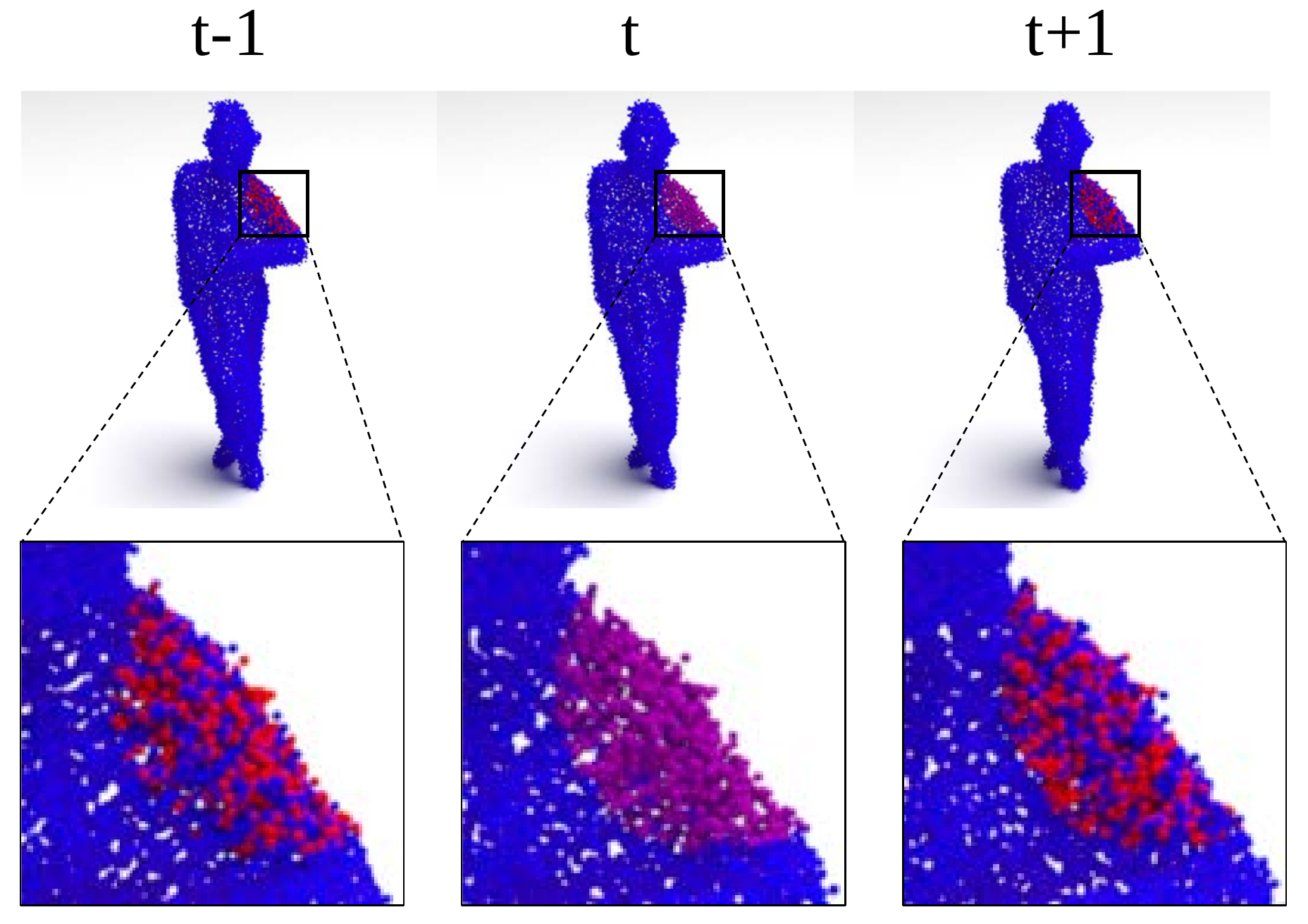}
\end{center}
\vspace{-0.1in}
  \caption{Visualization of the temporal correspondence search in adjacent frames. The patch in an adjacent frame that is collocated with the target patch in the $t$-th frame (colored in purple) is moved under the action of the gradient field in adjacent frames, until it reaches a balanced state (colored in red). }
\label{fig:corr_visualize}
\end{figure}

\subsubsection{Visualization of the temporal correspondence}
Fig.~\ref{fig:corr_visualize} illustrates the temporal correspondence search results by the proposed method. 
The target patch is colored purple while the searched corresponding patches are colored red. 
We see that the corresponding patches properly match the underlying surfaces in adjacent frames. 
This leads to the accurate estimation of the temporal gradient field for denoising. 
Also, as the temporal correspondence search is a fundamental problem in dynamic point cloud tasks, our correspondence search approach is applicable in other tasks such as motion estimation in dynamic point cloud compression and analysis. 

\subsection{Ablation studies}
For a better understanding of our method, we conduct two ablation studies: 1) evaluate the importance of the temporal correspondence; 2) evaluate the effectiveness of our temporal gradient field estimation.

\subsubsection{With/Without the temporal correspondence} 
As shown in the last row and the first row of Table~\ref{tab:ablation}, we compare the denoising performance of our method with and without the temporal correspondence on the MPEG dataset. 
In the version of "Ours-NoTemporal", we denoise each frame of the input dynamic point cloud independently, with the gradient field calculated based on the spatial information via Eq.~\ref{eq:ensemble}. 
We see that the temporal correspondence significantly improves the denoising performance at the high noise level. 
Even if the noise level is low, the performance is also improved by introducing the temporal correspondence.
This validates the importance of the temporal correspondence for dynamic point cloud denoising.

\subsubsection{Temporal gradient field estimation} 
We evaluate different estimation approaches for the temporal gradient field. 
In particular, we consider two intuitive ways to estimate the temporal gradient for comparison: 
a) The "Ours-mean" variant directly takes the average of the initial gradient fields of three adjacent frames, \textit{i.e.}, changing Eq.~\ref{eq:final-field} to $\bar\gG_{t}(\vx) = \frac{1}{3} \bigg(\gG_{t}(\vx) + \gG_{t-1}(\vx)  + \gG_{t+1}(\vx)\bigg)$ without any transformation.  
b) The "Ours-icp" variant adopts the widely-used Iterative Closest Point (ICP) algorithm \cite{Chetverikov02} to search for corresponding patches across frames, \textit{i.e.}, changing the gradient-based temporal correspondence search method in Section~\ref{sec:corr} to the ICP.
The "Ours-gradient" version is the proposed method.

As shown in the last three rows of Table~\ref{tab:ablation}, we compare the denoising performance of different temporal gradient field estimation approaches. 
The proposed "Ours-gradient" achieves the best performance at both low and high noise levels, with significant improvement especially at the high noise level. 
The performance of the "Ours-mean" variant degrades due to the temporal movement, which means averaging the gradient field directly will introduce noise. 
Besides, the ICP variant performs worse when the noise level is high. 
This is probably because large noise causes severe deformation of surface patches, which makes it difficult for the ICP algorithm to accurately register corresponding patches.

\begin{table}[t]
    \centering
    \begin{tabular}{l|ccc|ccc}
\toprule
\multicolumn{1}{c|}{Noise}                          & \multicolumn{3}{c|}{0.60\%}                       & \multicolumn{3}{c}{3.00\%}                       \\
 & CD             & HD             & P2M            & CD             & HD             & P2M            \\
\midrule
Ours-NoTmeporal                  & 0.628          & 0.037          & 0.104          & 4.829          & 1.439          & 3.690          \\
\midrule
Ours-mean                  & 0.632          & 0.037          & 0.107          & 4.679          & 1.329          & 3.518          \\
Ours-icp                   & 0.623          & 0.037          & 0.100          & 4.829          & 1.455          & 3.740          \\
Ours-gradient              & \textbf{0.621} & \textbf{0.036} & \textbf{0.099} & \textbf{4.442} & \textbf{1.236} & \textbf{3.329} \\
\bottomrule
\end{tabular}
    \caption{Ablation study of different temporal gradient field estimation methods under Gaussian noise. 
    CD is multiplied by $10^4$, HD is multiplied by $10^2$ and P2M is multiplied by $10^4$. 
    }
    \label{tab:ablation}
\end{table}



\section{Conclusion}
\label{sec:conclusion}
In this paper, we propose a dynamic point cloud denoising method based on gradient fields, exploiting the temporal correspondence among adjacent frames. 
In particular, leveraging on rigid motions in classical mechanics, we move each surface patch of the target point cloud frame in the gradient field of adjacent frames until reaching a balanced state to acquire the temporal correspondence. 
The temporal gradient is then estimated from the inversely transformed gradients of consecutive frames, which is adopted in the gradient ascent process to perform denoising. 
Experimental results demonstrate the effectiveness of our method over both synthetic noise and simulated LiDAR noise. 
Future works include applying the proposed temporal correspondence search method to other tasks of dynamic point clouds, such as compression and analysis. 


\bibliographystyle{IEEEtran}
\bibliography{egbib}

\begin{thebibliography}{10}
\providecommand{\url}[1]{#1}
\csname url@samestyle\endcsname
\providecommand{\newblock}{\relax}
\providecommand{\bibinfo}[2]{#2}
\providecommand{\BIBentrySTDinterwordspacing}{\spaceskip=0pt\relax}
\providecommand{\BIBentryALTinterwordstretchfactor}{4}
\providecommand{\BIBentryALTinterwordspacing}{\spaceskip=\fontdimen2\font plus
\BIBentryALTinterwordstretchfactor\fontdimen3\font minus
  \fontdimen4\font\relax}
\providecommand{\BIBforeignlanguage}[2]{{%
\expandafter\ifx\csname l@#1\endcsname\relax
\typeout{** WARNING: IEEEtran.bst: No hyphenation pattern has been}%
\typeout{** loaded for the language `#1'. Using the pattern for}%
\typeout{** the default language instead.}%
\else
\language=\csname l@#1\endcsname
\fi
#2}}
\providecommand{\BIBdecl}{\relax}
\BIBdecl

\bibitem{Alexa2003Computing}
M.~Alexa, J.~Behr, D.~Cohen-Or, S.~Fleishman, D.~Levin, and C.~T. Silva,
  ``Computing and rendering point set surfaces,'' \emph{IEEE Transactions on
  Visualization and Computer Graphics}, vol.~9, no.~1, pp. 0--15, 2003.

\bibitem{A2009Feature}
A.~C. {\"O}ztireli, G.~Guennebaud, and M.~Gross, ``Feature preserving point set
  surfaces based on non-linear kernel regression,'' in \emph{Computer Graphics
  Forum}, vol.~28, no.~2.\hskip 1em plus 0.5em minus 0.4em\relax Wiley Online
  Library, 2009, pp. 493--501.

\bibitem{Lipman2007Parameterization}
Y.~Lipman, D.~Cohen-Or, D.~Levin, and H.~Tal-Ezer, ``Parameterization-free
  projection for geometry reconstruction,'' \emph{ACM Transactions on
  Graphics}, vol.~26, no.~3, p.~22, 2007.

\bibitem{Hui2009Consolidation}
H.~Hui, L.~Dan, Z.~Hao, U.~Ascher, and D.~Cohen-Or, ``Consolidation of
  unorganized point clouds for surface reconstruction,'' \emph{ACM Transactions
  on Graphics (TOG)}, vol.~28, no.~5, pp. 1--7, 2009.

\bibitem{Huang2013Edge}
H.~Huang, S.~Wu, M.~Gong, D.~Cohen-Or, and H.~Zhang, ``Edge-aware point set
  resampling,'' \emph{ACM Transactions on Graphics (TOG)}, vol.~32, no.~1, pp.
  1--12, 2013.

\bibitem{mattei2017MRPCA}
E.~Mattei and A.~Castrodad, ``Point cloud denoising via moving rpca,'' in
  \emph{Computer Graphics Forum}, vol.~36, no.~8.\hskip 1em plus 0.5em minus
  0.4em\relax Wiley Online Library, 2017, pp. 123--137.

\bibitem{hu2020featuregraph}
W.~Hu, X.~Gao, G.~Cheung, and Z.~Guo, ``Feature graph learning for 3{D} point
  cloud denoising,'' \emph{IEEE Transactions on Signal Processing}, vol.~68,
  pp. 2841--2856, 2020.

\bibitem{sch2015graphbased}
Y.~Schoenenberger, J.~Paratte, and P.~Vandergheynst, ``Graph-based denoising
  for time-varying point clouds,'' in \emph{2015 3DTV-Conference: The True
  Vision-Capture, Transmission and Display of 3D Video (3DTV-CON)}.\hskip 1em
  plus 0.5em minus 0.4em\relax IEEE, 2015, pp. 1--4.

\bibitem{zeng2019GLR}
J.~Zeng, G.~Cheung, M.~Ng, J.~Pang, and C.~Yang, ``3d point cloud denoising
  using graph laplacian regularization of a low dimensional manifold model,''
  \emph{IEEE Transactions on Image Processing}, vol.~29, pp. 3474--3489, 2020.

\bibitem{qi2017pointnet}
C.~R. Qi, H.~Su, K.~Mo, and L.~J. Guibas, ``Pointnet: Deep learning on point
  sets for 3d classification and segmentation,'' in \emph{Proceedings of the
  IEEE conference on computer vision and pattern recognition}, 2017, pp.
  652--660.

\bibitem{qi2017pointnet2}
C.~R. Qi, L.~Yi, H.~Su, and L.~J. Guibas, ``Pointnet++: Deep hierarchical
  feature learning on point sets in a metric space,'' in \emph{Advances in
  neural information processing systems}, 2017, pp. 5099--5108.

\bibitem{wang2019dynamic}
Y.~Wang, Y.~Sun, Z.~Liu, S.~E. Sarma, M.~M. Bronstein, and J.~M. Solomon,
  ``Dynamic graph cnn for learning on point clouds,'' \emph{ACM Transactions on
  Graphics (TOG)}, vol.~38, no.~5, pp. 1--12, 2019.

\bibitem{hermosilla2019TotalDenoising}
P.~Hermosilla, T.~Ritschel, and T.~Ropinski, ``Total denoising: Unsupervised
  learning of 3d point cloud cleaning,'' in \emph{Proceedings of the IEEE
  International Conference on Computer Vision}, 2019, pp. 52--60.

\bibitem{duan2019NeuralProj}
C.~Duan, S.~Chen, and J.~Kovacevic, ``3d point cloud denoising via deep neural
  network based local surface estimation,'' in \emph{ICASSP 2019-2019 IEEE
  International Conference on Acoustics, Speech and Signal Processing
  (ICASSP)}.\hskip 1em plus 0.5em minus 0.4em\relax IEEE, 2019, pp. 8553--8557.

\bibitem{rakotosaona2020PCN}
M.-J. Rakotosaona, V.~La~Barbera, P.~Guerrero, N.~J. Mitra, and M.~Ovsjanikov,
  ``Pointcleannet: Learning to denoise and remove outliers from dense point
  clouds,'' in \emph{Computer Graphics Forum}, vol.~39, no.~1.\hskip 1em plus
  0.5em minus 0.4em\relax Wiley Online Library, 2020, pp. 185--203.

\bibitem{pistilli2020learning}
F.~Pistilli, G.~Fracastoro, D.~Valsesia, and E.~Magli, ``Learning
  graph-convolutional representations for point cloud denoising,'' \emph{arXiv
  preprint arXiv:2007.02578}, 2020.

\bibitem{luo2020DMR}
S.~Luo and W.~Hu, ``Differentiable manifold reconstruction for point cloud
  denoising,'' in \emph{Proceedings of the 28th ACM International Conference on
  Multimedia}, 2020, pp. 1330--1338.

\bibitem{luo2021score}
------, ``Score-based point cloud denoising,'' in \emph{Proceedings of the
  IEEE/CVF International Conference on Computer Vision}, 2021, pp. 4583--4592.

\bibitem{chen2021deep}
H.~Chen, B.~Du, S.~Luo, and W.~Hu, ``Deep point set resampling via gradient
  fields,'' \emph{arXiv preprint arXiv:2111.02045}, 2021.

\bibitem{arvanitis2018outliers}
G.~Arvanitis, A.~Spathis-Papadiotis, A.~S. Lalos, K.~Moustakas, and
  N.~Fakotakis, ``Outliers removal and consolidation of dynamic point cloud,''
  in \emph{IEEE International Conference on Image Processing (ICIP)}, 2018, pp.
  3888--3892.

\bibitem{buades2005non}
A.~Buades, B.~Coll, and J.-M. Morel, ``A non-local algorithm for image
  denoising,'' in \emph{IEEE Conference on Computer Vision and Pattern
  Recognition (CVPR)}, vol.~2, 2005, pp. 60--65.

\bibitem{dabov2007image}
K.~Dabov, A.~Foi, V.~Katkovnik, and K.~Egiazarian, ``Image denoising by sparse
  3-d transform-domain collaborative filtering,'' \emph{IEEE Transactions on
  image processing}, vol.~16, no.~8, pp. 2080--2095, 2007.

\bibitem{digne2012similarity}
J.~Digne, ``Similarity based filtering of point clouds,'' in \emph{IEEE
  Conference on Computer Vision and Pattern Recognition Workshops (CVPRW)},
  2012, pp. 73--79.

\bibitem{rosman2013patch}
G.~Rosman, A.~Dubrovina, and R.~Kimmel, ``Patch-collaborative spectral
  point-cloud denoising,'' in \emph{Computer Graphics Forum}, vol.~32,
  no.~8.\hskip 1em plus 0.5em minus 0.4em\relax Wiley Online Library, 2013, pp.
  1--12.

\bibitem{deschaud10}
J.~E. Deschaud and F.~Goulette, ``Point cloud non local denoising using local
  surface descriptor similarity,'' \emph{International Archives of
  Photogrammetry and Remote Sensing (IAPRS)}, vol.~38, pp. 109--114, 2010.

\bibitem{sarkar2018structured}
K.~Sarkar, F.~Bernard, K.~Varanasi, C.~Theobalt, and D.~Stricker, ``Structured
  low-rank matrix factorization for point-cloud denoising,'' in
  \emph{International Conference on 3D Vision (3DV)}, 2018, pp. 444--453.

\bibitem{zhou2021point}
Y.~Zhou, R.~Chen, Y.~Zhao, X.~Ai, and G.~Zhou, ``Point cloud denoising using
  non-local collaborative projections,'' \emph{Pattern Recognition}, vol. 120,
  p. 108128, 2021.

\bibitem{hu2021graph}
W.~Hu, J.~Pang, X.~Liu, D.~Tian, C.-W. Lin, and A.~Vetro, ``Graph signal
  processing for geometric data and beyond: Theory and applications,''
  \emph{IEEE Transactions on Multimedia}, 2021.

\bibitem{zeng20183d}
J.~Zeng, G.~Cheung, M.~Ng, J.~Pang, and Y.~Cheng, ``3{D} point cloud denoising
  using graph {Laplacian} regularization of a low dimensional manifold model,''
  \emph{IEEE Trans. Image Process.}, vol.~29, pp. 3474--3489, December 2019.

\bibitem{gao2018graph}
X.~Gao, W.~Hu, and Z.~Guo, ``Graph-based point cloud denoising,'' in \emph{IEEE
  Fourth International Conference on Multimedia Big Data (BigMM)}, 2018, pp.
  1--6.

\bibitem{duan2018weighted}
C.~Duan, S.~Chen, and J.~Kovacevic, ``Weighted multi-projection: 3d point cloud
  denoising with estimated tangent planes,'' \emph{arXiv preprint
  arXiv:1807.00253}, 2018.

\bibitem{irfan20213d}
M.~A. Irfan and E.~Magli, ``3d point cloud denoising using a joint geometry and
  color k-nn graph,'' in \emph{2020 28th European Signal Processing Conference
  (EUSIPCO)}.\hskip 1em plus 0.5em minus 0.4em\relax IEEE, 2021, pp. 585--589.

\bibitem{irfan2021joint}
------, ``Joint geometry and color point cloud denoising based on graph
  wavelets,'' \emph{IEEE Access}, vol.~9, pp. 21\,149--21\,166, 2021.

\bibitem{hu2021dynamic}
W.~Hu, Q.~Hu, Z.~Wang, and X.~Gao, ``Dynamic point cloud denoising via
  manifold-to-manifold distance,'' \emph{IEEE Transactions on Image
  Processing}, vol.~30, pp. 6168--6183, 2021.

\bibitem{lecun2006tutorial}
Y.~LeCun, S.~Chopra, R.~Hadsell, M.~Ranzato, and F.~Huang, ``A tutorial on
  energy-based learning,'' \emph{Predicting structured data}, vol.~1, no.~0,
  2006.

\bibitem{hyvarinen2005estimation}
A.~Hyv{\"a}rinen, ``Estimation of non-normalized statistical models by score
  matching,'' \emph{Journal of Machine Learning Research}, vol.~6, no. Apr, pp.
  695--709, 2005.

\bibitem{song2019generative}
Y.~Song and S.~Ermon, ``Generative modeling by estimating gradients of the data
  distribution,'' in \emph{Advances in Neural Information Processing Systems},
  2019, pp. 11\,918--11\,930.

\bibitem{luo2020differentiable}
S.~Luo and W.~Hu, ``Differentiable manifold reconstruction for point cloud
  denoising,'' in \emph{Proc. ACM Int. Conf. Multimedia}, October 2020, pp.
  1330--1338.

\bibitem{huang2017densely}
G.~Huang, Z.~Liu, L.~Van Der~Maaten, and K.~Q. Weinberger, ``Densely connected
  convolutional networks,'' in \emph{Proceedings of the IEEE conference on
  computer vision and pattern recognition}, 2017, pp. 4700--4708.

\bibitem{liu2019densepoint}
Y.~Liu, B.~Fan, G.~Meng, J.~Lu, S.~Xiang, and C.~Pan, ``Densepoint: Learning
  densely contextual representation for efficient point cloud processing,'' in
  \emph{Proceedings of the IEEE International Conference on Computer Vision},
  2019, pp. 5239--5248.

\bibitem{fps}
P.~Kamousi, S.~Lazard, A.~Maheshwari, and S.~Wuhrer, ``Analysis of farthest
  point sampling for approximating geodesics in a graph,'' \emph{Computational
  Geometry}, vol.~57, pp. 1--7, 2016.

\bibitem{MPEG}
T.~Ebner, I.~Feldmann, O.~Schreer, P.~Kauff, and T.~Unger, ``Hhi point cloud
  dataset of a boxing trainer,'' in \emph{ISO/IEC JTC1/SC29/WG11 (MPEG2018)
  input document M42921}, July 2018.

\bibitem{Cai16}
C.~Loop, Q.~Cai, S.~O. Escolano, and P.~A. Chou, ``Microsoft voxelized upper
  bodies - a voxelized point cloud dataset,'' in \emph{ISO/IEC JTC1/SC29 Joint
  WG11/WG1 (MPEG/JPEG) input document m38673/M72012}, May 2016.

\bibitem{digne2017bilateral}
J.~Digne and C.~De~Franchis, ``The bilateral filter for point clouds,''
  \emph{Image Processing On Line}, vol.~7, pp. 278--287, 2017.

\bibitem{cazals2005jetsfit}
F.~Cazals and M.~Pouget, ``Estimating differential quantities using polynomial
  fitting of osculating jets,'' \emph{Computer Aided Geometric Design},
  vol.~22, no.~2, pp. 121--146, 2005.

\bibitem{TotalDenoising2019}
P.~Hermosilla, T.~Ritschel, and T.~Ropinski, ``Total denoising: Unsupervised
  learning of 3d point cloud cleaning,'' in \emph{Proceedings of the IEEE
  International Conference on Computer Vision}, 2019, pp. 52--60.

\bibitem{fan2017pointsetgen}
H.~Fan, H.~Su, and L.~J. Guibas, ``A point set generation network for 3d object
  reconstruction from a single image,'' in \emph{Proceedings of the IEEE
  conference on computer vision and pattern recognition}, 2017, pp. 605--613.

\bibitem{Huttenlocher93}
D.~P. Huttenlocher, G.~A. Klanderman, and W.~J. Rucklidge, ``Comparing images
  using the hausdorff distance,'' \emph{IEEE Transactions on Pattern Analysis
  \& Machine Intelligence}, vol.~15, no.~9, pp. 850--863, 1993.

\bibitem{ravi2020pytorch3d}
N.~Ravi, J.~Reizenstein, D.~Novotny, T.~Gordon, W.-Y. Lo, J.~Johnson, and
  G.~Gkioxari, ``Accelerating 3d deep learning with pytorch3d,''
  \emph{arXiv:2007.08501}, 2020.

\bibitem{Meshlab}
P.~Cignoni, M.~Callieri, M.~Corsini, M.~Dellepiane, F.~Ganovelli, and
  G.~Ranzuglia, ``Meshlab: an open-source mesh processing tool,'' in
  \emph{Eurographics Italian Chapter Conference}, Salerno, Italy.\hskip 1em
  plus 0.5em minus 0.4em\relax The Eurographics Association, 2008, pp.
  129--136.

\bibitem{gschwandtner2011blensor}
M.~Gschwandtner, R.~Kwitt, A.~Uhl, and W.~Pree, ``Blensor: Blender sensor
  simulation toolbox,'' in \emph{International Symposium on Visual
  Computing}.\hskip 1em plus 0.5em minus 0.4em\relax Springer, 2011, pp.
  199--208.

\bibitem{Chetverikov02}
D.~Chetverikov, D.~Svirko, D.~Stepanov, and P.~Krsek, ``The trimmed iterative
  closest point algorithm,'' in \emph{International Conference on Pattern
  Recognition, 2002. Proceedings}, vol.~3, 2002, pp. 545--548.

\end{thebibliography}

\end{document}